\pdfoutput=1

\documentclass[11pt]{article}

\usepackage[]{ACL2023}

\usepackage{todonotes}

\usepackage{times}
\usepackage{latexsym}
\usepackage{tabularx}
\usepackage{graphicx} 
\usepackage[T1]{fontenc}

\usepackage[utf8]{inputenc}

\usepackage{microtype}

\usepackage{inconsolata}

\usepackage{booktabs}
\usepackage{url}
\usepackage{inconsolata}
\usepackage{threeparttable}
\usepackage{multirow}
\usepackage{xspace}
\usepackage{algorithm}
\usepackage{algpseudocode}

\usepackage{hyperref}
\usepackage{amsmath}

\usepackage{setspace}

\usepackage{makecell}

\usepackage{array}
\usepackage{arydshln}

\usepackage{rotating}

\usepackage{subcaption}
\usepackage[most]{tcolorbox}

\newcommand{\frameworkname}{\textsc{LLMCrit}\xspace}
\definecolor{weizhey}{rgb}{0.43, 0.71, 0.40}

\definecolor{myblue}{rgb}{0.9, 0.1, 0.94}

\definecolor{stackgreen}{rgb}{0.55294,0.74118,0.52157}
\definecolor{stackred}{rgb}{1,0.22549,0.21373}

\newenvironment{enumerate*}%
 {\leftmargini=10pt\begin{enumerate}%
  \setlength{\itemsep}{0pt}%
  \setlength{\parskip}{0pt}}%
 {\end{enumerate}}

\newenvironment{itemize*}%
 {\leftmargini=10pt\begin{itemize}%
  \setlength{\itemsep}{0pt}%
  \setlength{\parskip}{0pt}%
  }%
 {\end{itemize}}

\title{Knowledge-based Feedback Generation}
\title{\frameworkname : Teaching Large Language Models to Use Criteria}

\author{Weizhe Yuan \\
  New York University \\
  \texttt{wy885@nyu.edu} \\\And
  Pengfei Liu \\
  Shanghai Jiao Tong University \\
  \\\And
  Matthias Gallé \\
  Cohere\\ 
  \\}
\begin{document}
\maketitle
\begin{abstract}

Humans follow \textit{criteria} when they execute tasks, and these criteria are directly used to assess the quality of task completion. Therefore, having models learn to use criteria to provide feedback can help humans or models to perform tasks better. However, current research in this area tends to consider only a limited number of criteria, or only a limited number of quality assessment aspects. To fill this gap,  we propose a general framework that enables large language models (LLMs) to use comprehensive criteria for a task in delivering natural language feedback on task execution. In particular, we present a \textit{model-in-the-loop} framework that semi-automatically derives criteria from collected guidelines for different writing tasks and constructs in-context demonstrations for each criterion. We choose three tasks from real-world scenarios to operationalize this idea: paper introduction writing, Python code writing, and Reddit post writing, and evaluate our feedback generation framework using different LLMs. The results reveal the fine-grained effects of adding criteria and demonstrations and provide valuable guidance on how to teach LLMs to use criteria more effectively.

\end{abstract}

\section{Introduction}
\begin{figure}[t]
    \centering
    \begin{subfigure}{\linewidth}
        \centering
        \includegraphics[width=\textwidth]{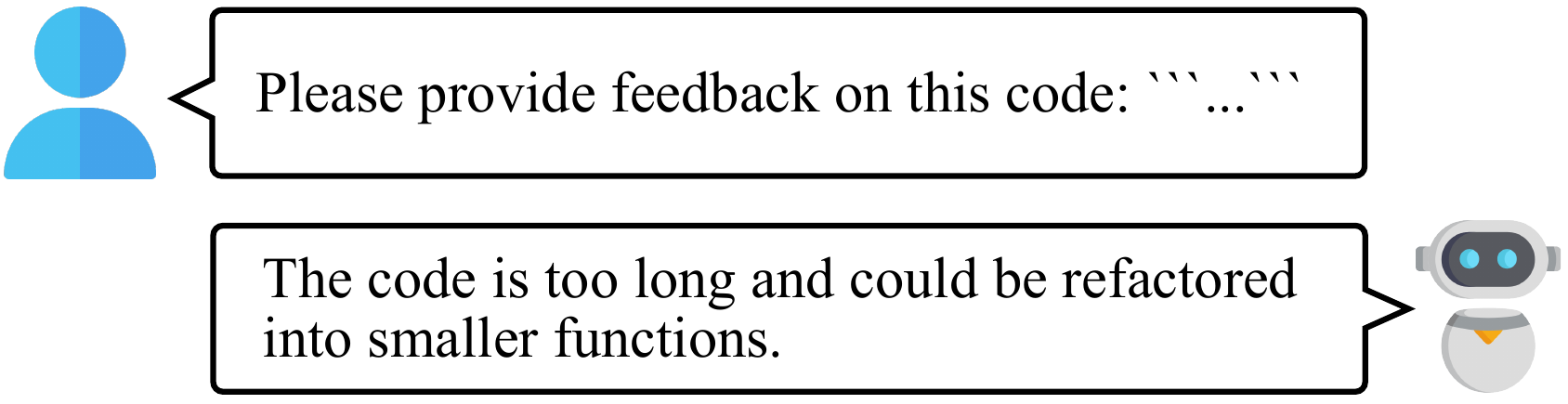}
        \caption{Original feedback generation.}
        \label{subfig:a}
    \end{subfigure}

    \begin{subfigure}{\linewidth}
        \centering
        \includegraphics[width=\textwidth]{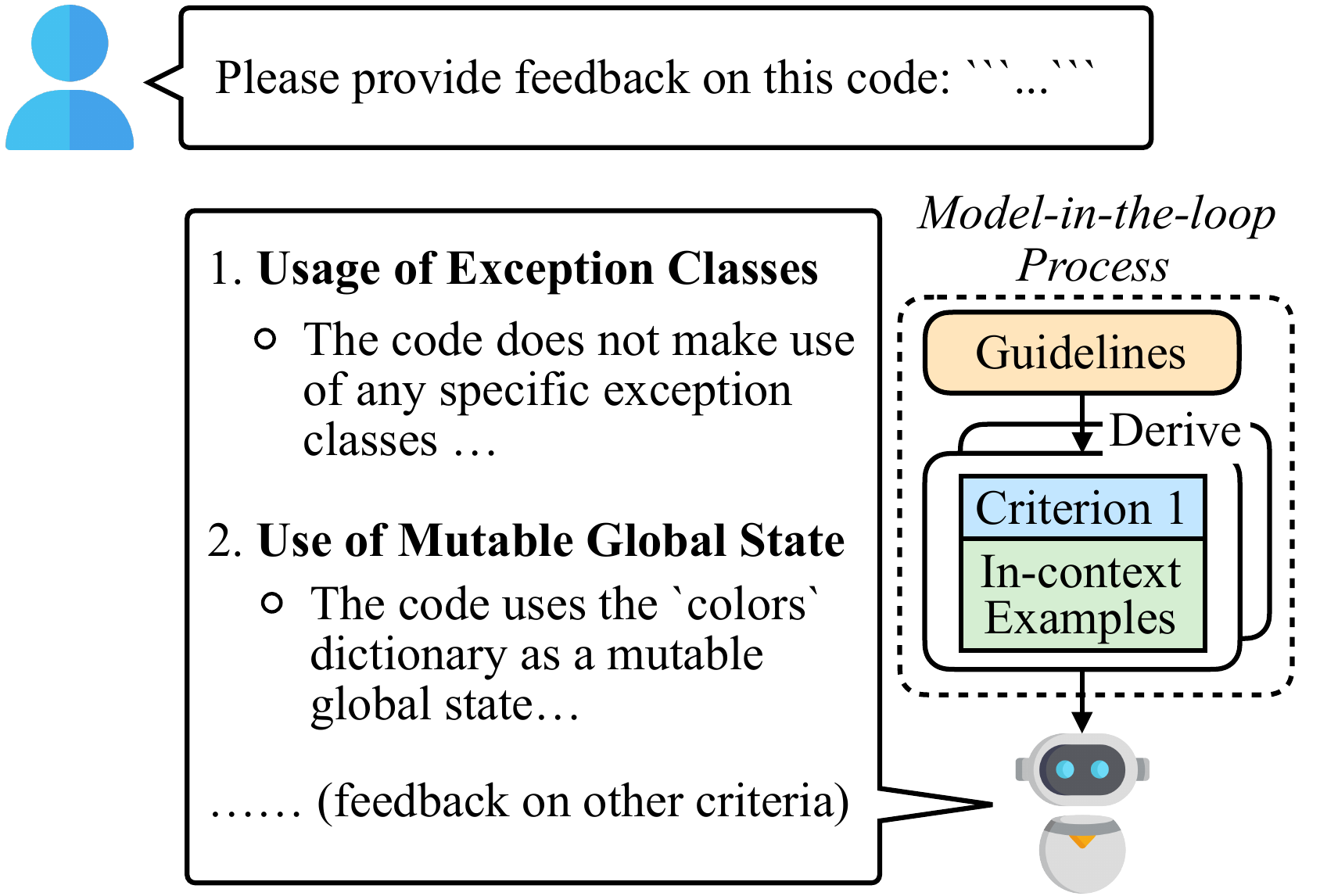}
        \caption{Teach LLM to use criteria for feedback generation.}
        \label{subfig:b}
    \end{subfigure}
    \caption{Illustration of teaching LLMs to use criteria.}
    \label{fig:fig1}
\end{figure}
A \textit{criterion} is a principle or standard by which something may be judged or decided~\citep{criterion}. An example criterion for a writing task could be that ``the text should not contain grammar errors''. In the fields of cognitive science and psychology, it is widely recognized that humans, when engaging in a task, invariably follow certain criteria~\citep{Anderson1990TheAC} which are often directly used for judging the quality of task completion. Despite the importance of criteria, the current field of research has relatively neglected this piece of human wisdom, with works mainly focusing on using a limited number of criteria~\citep{bai2022constitutional, DBLP:journals/corr/abs-2305-03047, kundu2023specific} or using comprehensive criteria to evaluate specific aspects such as safety~\citep{xu2023align}. If we can teach LLMs to use comprehensive criteria to judge the quality of task completion from various aspects, on the one hand, this would greatly increase human productivity, as one can enter a cycle of rapid iterative improvement by incorporating feedback from LLMs. On the other hand, this may also be part of the scalable oversight~\citep{bowman2022measuring} solution. For even if future models are capable of superhuman performance, using criteria to judge the quality of task completion ensures that they align with human values and continue to improve.

However, there are challenges in teaching LLMs to use criteria: (1) Criteria are often times implicit in the human written guidelines (e.g., implicit in a book on how to write good academic papers) rather than in the form of an explicit list. (2) Even if we have access to a list of criteria, we may overlook or misapply some due to (a) the large number of them and the fact that the understanding of some of them (b) requires expertise.

In light of these challenges, we introduce the task of ``\textit{teaching large language models to use criteria}''. Specifically, we concentrate on instructing LLMs to generate natural language (NL) feedback on task execution based on comprehensive criteria. To address the two challenges we mentioned above, we propose a general framework \frameworkname that uses a model-in-the-loop process that semi-automatically leverages existing human-written guidelines for each task, as shown in Fig.~\ref{fig:fig1}. In particular, we first use an LLM to extract criteria from the guidelines to solve challenge 1. To solve challenge 2-(b), we use an LLM to construct demonstrations for each criterion to teach the model how to determine whether the execution of a task meets a certain criterion inspired by studies that have taught LLMs to use tools through prompting \citep{DBLP:journals/corr/abs-2305-17126, shen2023hugginggpt, DBLP:journals/corr/abs-2303-16434}. To address challenge 2-(a), we explored the effect of providing criteria to LLMs at different granularities (including giving the model one criterion at a time, or a set of related criteria at a time).

In order to get a clearer understanding of whether the generated feedback is helpful or not, and what makes it less helpful, we propose layered evaluation metrics, where the quality of the feedback is evaluated in a hierarchical manner from four perspectives: \textit{validity}, \textit{contextualization}, \textit{constructiveness}, and \textit{helpfulness}. We assess our framework using three real-world writing tasks: scientific paper writing, Python code development, and Reddit post creation. Experiment results show how criteria and demonstrations impact the quality of generated feedback from four perspectives, and how providing criteria at different granularities affects the quality of generated feedback. These results provide valuable guidance for us to teach LLMs to use criteria in the future. Our contributions are:

\begin{enumerate*}
\item We propose a new framework \frameworkname for obtaining scalable oversight that takes advantage of criteria. \frameworkname starts with existing guidelines, semi-automatically extracts criteria from it, and constructs demonstrations for each criterion. We then apply these criteria and demonstrations to guide LLMs to generate NL feedback.

\item We instantiate our framework with three writing tasks: scientific paper writing, Python code writing, and Reddit post writing. Meanwhile, to address the difficulty in evaluating feedback, we propose layered evaluation metrics to measure feedback quality, allowing for a more clean and organized assessment.

\item Experiment results suggests that providing criteria allows the model to generate feedback that contains more critiques and suggestions. In addition, providing demonstrations makes the resulting critiques and suggestions more helpful, but may also distract the model from generating feedback on the demonstration input.

\item We release 83 criteria and 332 in-context demonstrations that we have collected and curated for the three real-world writing tasks to the community for future research. All resources can be found at \url{https://github.com/yyy-Apple/LLMCrit}.


\end{enumerate*}

\section{Related Work}




\paragraph{Augmented LLMs}

Augmenting LLMs with additional context has become a powerful tool for continual learning as well as combating hallucinations. 
These augmentations include learned rule library \citep{zhu2023large} to perform deductive reasoning, retrievers to retrieve relevant documents from knowledge bases to answer fact-related queries \citep{DBLP:conf/nips/LewisPPPKGKLYR020, trivedi-etal-2023-interleaving} or various tools and APIs to perform specialized tasks. \citep{shen2023hugginggpt, DBLP:journals/corr/abs-2303-16434}. 
We continue this line of work by proposing to augment LLMs with \textit{criteria} extracted from the writing task guidelines so that models can generate more helpful feedback for a piece of writing for that task.

\paragraph{Automated Critique Generation} 
It is commonplace to use LLMs as \textit{evaluators}~\citep{alpaca_eval, DBLP:journals/corr/abs-2306-05685,li2023generative,pandalm2024} -- even replacing human evaluations for training~\cite{cui2023ultrafeedback} --  which can be seen as a basic way of critique.
Having models criticize their own output and then rewriting is another popular use of synthetic data creation~\cite{bai2022constitutional,madaan2023selfrefine,wang2023shepherd}. Departing from that, our focus is on teaching LLMs to learn criteria so that they can generate more comprehensive feedback that is aligned with human values.

\begin{figure*}
    \centering
    \includegraphics[width=0.98\linewidth]{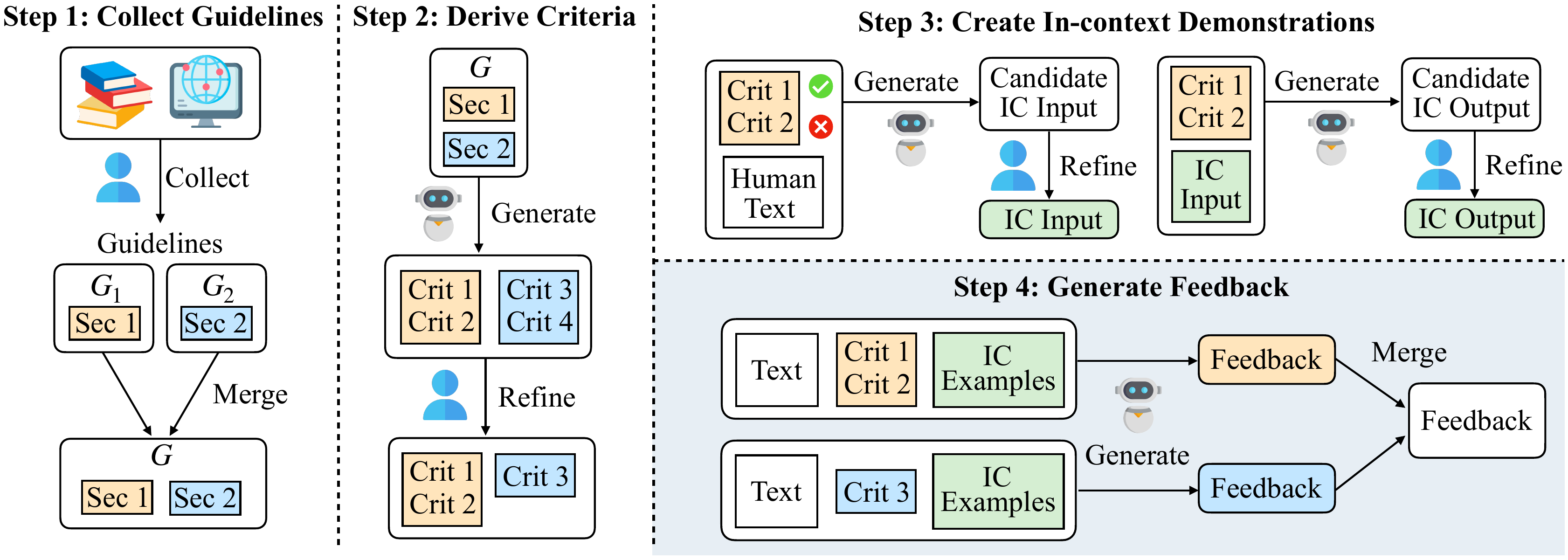}
    \caption{Our \frameworkname framework for teaching LLMs to use criteria. By applying a model-in-the-loop approach, we semi-automatically derive criteria and construct in-context demonstrations for each criterion. ``Sec'' stands for ``section'', ``Crit'' stands for ``criterion'', ``IC'' stands for ``in-context''. Step 1, 2, and 3 only need to be completed once, and the resulting criteria and demonstrations can be reused by different LLMs in Step 4 (shaded).
    }
    \label{fig:framework}
\end{figure*}

\begin{figure}[!t]
    \centering
    \includegraphics[width=0.75\linewidth]{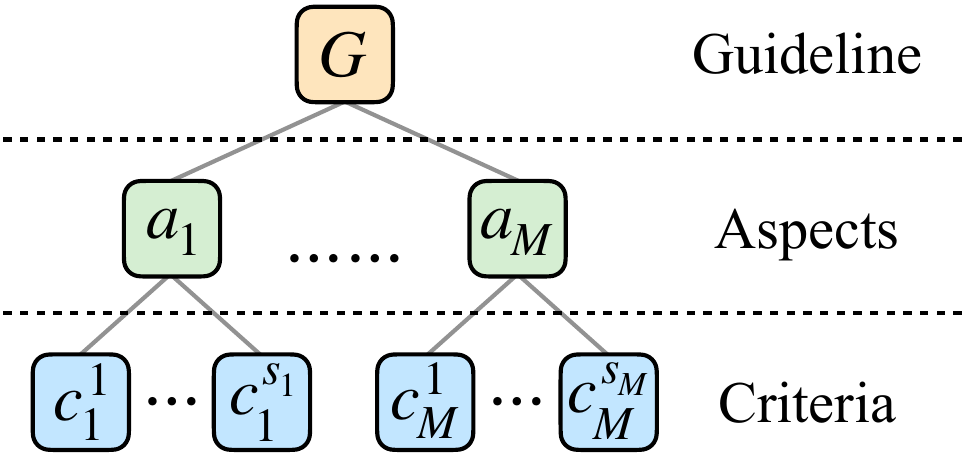}
    \caption{Hierarchical tree structure of the writing task guideline $G$.}
    \label{fig:hierarchy}
\end{figure}
\section{\frameworkname Framework}
Our proposed framework is shown in Fig.~\ref{fig:framework}. There are four steps within this framework. 


\subsection{Collect Guidelines}
For each task, we collect guidelines from authoritative sources, such as documents or books written by domain experts. 
These guidelines are not just arbitrary recommendations; rather, they are clear, structured, and comprehensive. 
This step entails the most manual intervention, but has to be done once for each domain.
As an example, for code we use official style guides while for social media posts we leverage community rules as well as well-regarded writing advice.

\subsection{Derive Criteria}
Upon gathering guidelines, we derive criteria from them. To achieve this, we apply a two-step approach whereby we first decompose the guidelines into different aspects (sections), and then further extract criteria for each aspect, as shown in Fig.~\ref{fig:hierarchy}.

\noindent\textbf{Aspects}
The derivation of aspects is mainly based on the sectional division within the guideline. Typically, a guideline contains various sections, each focusing on an aspect of writing. For example, in a Python style guide, there are sections related to exception handling and mutable global state.

\noindent\textbf{Criteria}
Upon segmenting the guidelines into different aspects, for each aspect, we extract a fixed set of criteria (e.g., for the mutable global state aspect of the Python style guide, we can extract fine-grained criteria about the naming conventions for mutable global entities, etc.). We use an LLM to interactively extract criteria for each aspect, followed by manual review to refine and remove duplicates.

\subsection{Create Demonstrations}
\label{subsec:create_demonstrations}
We provide criteria and demonstrations in the granularity of an aspect instead of a single criterion (in \S\ref{subsection:analysis}, we ablate the impact of providing criteria at different granularities). For each aspect $a_i$, the goal is to collect $k$ demonstrations, $(x_i^1, y_i^1)\cdots(x_i^k, y_i^k)$, where $x_i^j$ is the input text to be examined and $y_i^j$ is the corresponding feedback text.

\noindent\textbf{Create Demonstration Input}
To increase the diversity of demonstrations, we aim to collect input texts that randomly violate some criteria while complying with others in an aspect. For an aspect $a_i$ with criteria $c_i = \{c_i^1, \cdots, c_i^N\}$, we first sample a satisfaction vector $s_i = \{s_i^1, \cdots, s_i^N\}$ where $s_i^j=1$ indicates satisfaction of criterion $c_i^j$ and $s_i^j=0$ the violation of $c_i^j$. Then, we prompt an LLM with $a_i$, $c_i$, and $s_i$ to get a candidate input text $x_i$, followed by manual refinement to ensure compliance with $s_i$. We repeat this process $k$ times to collect $k$ demonstrations for $a_i$. 
One point worth noting is that our preliminary experiments suggest that by relying solely on $a_i$, $c_i$, and $s_i$ to create the input text, LLMs (even the most advanced ones) are likely to generate text that contains unnatural and simplistic errors. To address this problem, we use existing human-written texts $\tilde{x}_i^1\cdots\tilde{x}_i^k$ as a starting point. These texts are then modified by the LLM to introduce specific criteria violations.



\noindent\textbf{Create Demonstration Output}
To get the demonstration output $y_i^j$, we prompt an LLM with $x_i^j$, $a_i$ $c_i$, and $s_i^j$ to get the initial feedback. 
Subsequently, human experts would refine this output, addressing minor issues such as factual errors and enhancing its structure and clarity. 



\subsection{Generate Feedback}
This step differs from earlier ones because it is repeatable for any LLM of interest, utilizing the criteria and demonstrations obtained in previous steps. The feedback generation process for a specific LLM is an iterative process. Given the input writing $w$, we go through all the aspects, and for each aspect $a_i$, we prompt the LLM with $w$, $a_i$, $c_i$ and $(x_i^j, y_i^j)_{j=1\cdots k}$ to get its feedback text $f_i$. The final feedback obtained using the given LLM is a collection of feedback text for each aspect $\{f_1, \cdots, f_M\}$, where $M$ is the number of aspects. 

\section{Data and Evaluation Foundation}
\label{sec:data_foundataion_for_experiments}
To demonstrate the effectiveness of \frameworkname, we conduct experiments on three application scenarios: \textit{paper introduction writing}, \textit{Python code writing}, and \textit{Reddit post writing}. For data preparation including criteria extraction, demonstration creation, and test data construction, we used Claude2~\citep{claude2} prompting with the chain-of-thought (CoT) technique~\citep{NEURIPS2022_9d560961}.

\subsection{Paper Introduction Writing}

\paragraph{Criteria} We sourced our guidelines for scientific paper writing from the Scribbr website.\footnote{\url{https://www.scribbr.com/research-paper}} From this resource, we manually extracted specific aspects tailored for the introduction part of a scientific paper. As a result, we curated five aspects. We used the prompt in Appendix Tab.~\ref{tab:prompt_for_extracting_criteria} for criteria extraction, which resulted in 11 criteria in total. An example criterion of this task is ``\textit{\textbf{Introduce Your Topic}: Does the introduction effectively identify the subject matter and provide sufficient background to inform the reader about the topic being addressed?}''.

\noindent\textbf{Demonstrations} We crafted two demonstrations for each aspect. See Appendix Tab.~\ref{tab:prompt_for_ic_input_intro} for input creation prompt and Tab.~\ref{tab:prompt_for_ic_output} for output creation prompt.

\noindent\textbf{Test Data} We collected all ICLR accepted and rejected papers from 2020 to 2022. For each paper, we used the URL information to download the PDF version and used GROBID \citep{GROBID} to parse the content into XML format, which we could easily extract the introduction part. We randomly selected 50 accepted papers and 50 rejected papers so that we collected 100 introductions in total. 

\subsection{Python Code Writing}
\paragraph{Criteria} 
We collected our guidelines for Python code writing from the Google Python Style Guide.\footnote{\url{https://google.github.io/styleguide/pyguide.html}} We specifically focused on aspects that are not easily addressed by standard automated tools. Consequently, we selected a set of 14 distinct aspects. We used the prompt in Appendix Tab.~\ref{tab:prompt_for_extracting_criteria} for criteria extraction and obtained a total of 47 individual criteria for Python code writing. An example criterion of this task is ``\textit{\textbf{Usage of Exception Classes}: Verify whether the code makes appropriate use of built-in exception classes (e.g., `ValueError`, `TypeError`, etc.) instead of using generic exceptions or `assert` statements for public API argument validation.}''

\noindent\textbf{Demonstrations} We crafted four demonstrations for each aspect. See Appendix Tab.~\ref{tab:prompt_for_ic_input_code} for input creation prompt and Tab.~\ref{tab:prompt_for_ic_output} for output creation prompt.

\noindent\textbf{Test Data} We utilized the COMMITPACKFT dataset from \citet{muennighoff2023octopack} as the foundation for our data. To guarantee complexity within our test set, we filtered out Python files containing less than 30 lines of code, not counting comments. From this refined pool, we then randomly selected 100 files to comprise our test data.

\subsection{Reddit Post Writing}
We selected the WritingPrompts Subreddit as our target community, which is centered around text-based content.

\noindent\textbf{Criteria}
We collected our guidelines for WritingPrompts post composition from both the community's rules\footnote{\url{https://www.reddit.com/r/WritingPrompts/wiki/rules/}}\footnote{\url{https://www.reddit.com/r/WritingPrompts/wiki/prompts/}} and best practices for writing prompts, as well as from the general policies governing Reddit posts.\footnote{\url{https://www.redditinc.com/policies/content-policy}}\footnote{\url{https://support.reddithelp.com/hc/en-us/articles/205926439}} After curation, we obtained a set of six aspects. We used the prompt in Appendix Tab.~\ref{tab:prompt_for_extracting_criteria} for criteria extraction and got a total of 25 criteria. An example criterion of this task is ``\textit{\textbf{Advertising Prohibition}: Review the prompt to ascertain it's not a disguised advertisement, promotion, or any form of covert marketing.}''

\noindent\textbf{Demonstrations} We crafted four demonstrations for each aspect. See Appendix Tab.~\ref{tab:prompt_for_ic_input_reddit} for input creation prompt and Tab.~\ref{tab:prompt_for_ic_output} for output creation prompt.

\noindent\textbf{Test Data} We gathered data by crawling the WritingPrompts subreddit, collecting the 1,000 most popular submissions tagged with \texttt{[WP]}. From these, we randomly chose 100 posts. 

\subsection{Diversify Test Data}
After collecting 100 test data for each task, we further modified 50 of them to selectively violate some of the criteria through LLM prompting so that the data distribution could be more diverse. We applied the same method in \S\ref{subsec:create_demonstrations} for creating demonstration input, except that we did not manually refine the model-generated text.

\subsection{Evaluation Methodology}
The most straightforward way to judge the quality of the model-generated feedback is to measure whether it provides helpful (i.e., actionable and free of factual errors) critiques.\footnote{Throughout the paper, critiques refer to violations of the criteria, or suggestions how to improve the writing with respect to those criteria.} Given the resource-intensive nature of human evaluation and recent findings indicating that LLMs can assess model outputs in a manner that aligns with human evaluations~\citep{alpaca_eval, zheng2023judging}, we resort to model-based evaluation. We employ Claude2 for evaluation due to its free research accessibility and support for large context length. However, our initial findings indicate that Claude2 struggles with assessing feedback helpfulness in a single step, as its judgments often do not correlate well with human evaluations. Upon detailed examination of the feedback generated by LLMs, we identified several key issues that contribute to the unhelpfulness of the feedback. These issues can be categorized into four primary types of errors:
\begin{enumerate*}
\item \textbf{Validity Error}: The generated text does not qualify as feedback, but rather offers general writing advice, extends the given criterion, or simply repeats it.
\item \textbf{Context Error}: The feedback pertains to an in-context example rather than the specific example at hand.
\item \textbf{Insufficient Critique}: The feedback lacks depth and specificity. It may merely summarize the original text, provide broad and general comments without detailed critiques, or offer only positive remarks without any constructive suggestions.
\item \textbf{Unhelpful Critique}: The feedback includes critiques or suggestions that are either too vague to be actionable or are factually incorrect
\end{enumerate*}
In light of these challenges, we propose a refined evaluation methodology. We suggest implementing a layered evaluation strategy that breaks down the assessment of ``helpfulness'' into the following four progressive perspectives. 

\begin{enumerate*}
    \item \textbf{Validity} measures whether the generated text is a valid feedback text. 
\item \textbf{Contextualization} measures whether the generated text is a feedback text specific to the current input. 
\item \textbf{Constructiveness} measures whether the feedback text provides critiques or suggestions w.r.t. a certain criterion. 
\item \textbf{Helpfulness} measures whether the feedback text is helpful for improving the text writing w.r.t. a certain criterion. 
\end{enumerate*}

By sequentially applying these perspectives, we can efficiently filter out feedback that fails to meet lower-level perspectives, ensuring that only the most relevant and constructive feedback reaches the final stage of evaluation. This method significantly reduces the noise in the feedback that is ultimately assessed for helpfulness. We designed different prompt templates for evaluating each perspective (see Appendix Tab.~\ref{tab:prompt_for_evaluating_feedback_validity_intro}\textasciitilde Tab.~\ref{tab:prompt_for_evaluating_feedback_constructiveness_helpfulness_reddit_part2}) and validated our model-based evaluation approach by running a small-scale human evaluation conducted by two experienced NLP researchers using the same prompts as annotation guidelines. For each perspective and task, we randomly chose 100 feedback samples from all models, totaling 300 samples per perspective for annotation. The results indicate that our model-based evaluation method is highly accurate, with average accuracies of 96.48\% for validity (95.33\% inter-annotator agreement), 94.23\% for contextualization (92\% inter-annotator agreement), 95.22\% for constructiveness (91\% inter-annotator agreement), and 88.53\% for helpfulness (89.67\% inter-annotator agreement).\footnote{When calculating the accuracy, we only include the samples where both annotators agreed on the label.} More details are in Appendix~\ref{app:meta-eval}.

\section{Experiments}
\label{sec:experiments}
To investigate the best way of teaching LLMs to use criteria, we selected leading open-source and proprietary models for our experiments. We specifically choose models with a minimum context length of 8k tokens to ensure they can accommodate the demonstrations within the prompts.

\paragraph{Models}
For pure text understanding tasks, we consider the following LLMs: Command-52b~\citep{command}, GPT4~\citep{OpenAI_GPT4_2023}, Together-7b~\citep{together7b}, and LongAlpaca-13b~\citep{long-alpaca}. For tasks involving code understanding, we select the following LLMs that have been trained on code data\footnote{According to our preliminary experiments, most text models do not do well on code tasks.}: GPT4, Claude2~\citep{claude2}, CodeLLAMA-13b~\citep{rozière2023code} and WizardCoder-13b~\citep{luo2023wizardcoder}.

\paragraph{Teaching Strategies}
We consider four strategies based on whether we provide criteria and demonstrations to the model. The exact prompts are in Appendix 
Tab.~\ref{tab:prompt_for_generating_feedback_intro}\textasciitilde Tab.~\ref{tab:prompt_for_generating_feedback_reddit}.

\begin{itemize*}
    \item \textbf{Base}:  We do not provide criteria nor demonstrations.
    \item \textbf{Crit}: We only provide the guideline for each aspect and the extracted criteria. 
    \item \textbf{ICL}: We only provide demonstrations, formatted in a structured way.
    \item \textbf{CrICL}: We provide both the aspect guideline, extracted criteria, and demonstrations.
\end{itemize*}

In the case of the Crit, ICL, and CrICL strategies, the resulting feedback text is a collection of feedback text for each of the aspects.

\paragraph{Decoding Strategy}
In obtaining the feedback text for the base strategy, and in obtaining the feedback text for one aspect of the other three strategies, we use nucleus sampling~\citep{conf/iclr/HoltzmanBDFC20} with temperature $T=0.5$, $p=1.0$ and sample five generations. Then we apply self-consistency techniques from~\citet{jain2023selfconsistency} to select the optimal one based on its cosine similarity with all other generations. We use Cohere \texttt{embed-english-v2.0} model to embed sentences when calculating cosine similarity between generations.

\paragraph{Human Annotation Details}
Both processes of deriving criteria and creating demonstrations  require meticulous manual refinement. We provide details of this process involving human input below.
For this study, this has been carried out by two experienced NLP researchers.
In terms of the human annotators' role in verifying the collection of criteria, their main responsibility is to identify and remove duplicates without making additional edits. With a dataset comprising three tasks, 25 aspects, and 83 criteria, the average number of criteria per aspect is approximately 3.3. The time required to review each criterion for duplication is about 10 seconds, resulting in a total deduplication time of roughly 30 minutes.
For the creation of demonstration inputs and outputs, our approach involved 332 demonstrations across three tasks. On average, each demonstration required about 8 minutes of curation and editing for both the input and the output. This effort amounted to a total of approximately 44 hours of human annotation.

\section{Results}
\label{subsec:results}
\subsection{Validity and Contextualization}
\label{subsubsec:validity_and_contextualization}
\begin{table*}[!ht]
\centering
\footnotesize
 \setlength\tabcolsep{4pt}
\begin{tabular}{ccc}
\toprule

& \textbf{Constructiveness} &\\
\midrule
\multirow{13}[1]{*}{\includegraphics[scale=0.37]{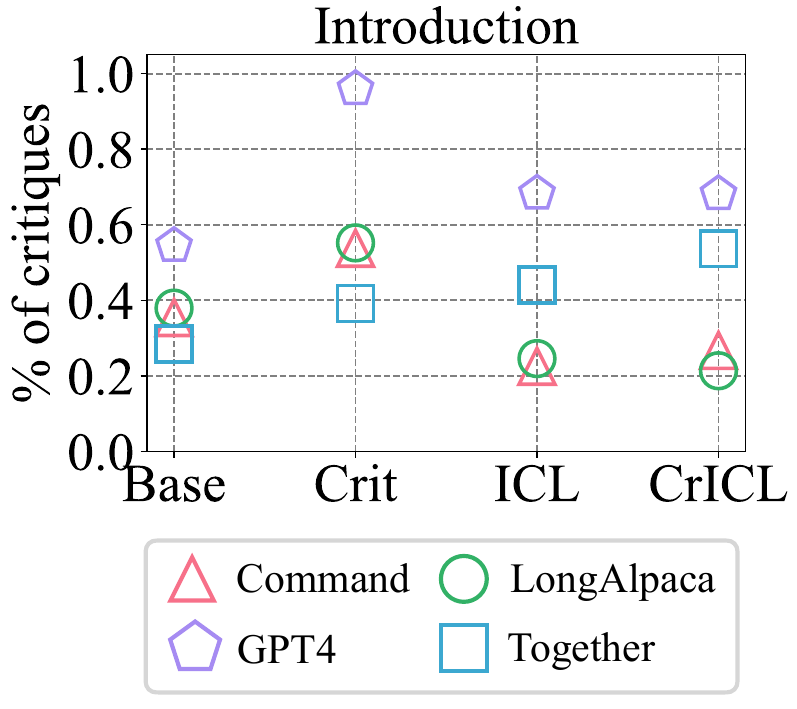}}  & \multirow{13}[1]{*}{\includegraphics[scale=0.37]{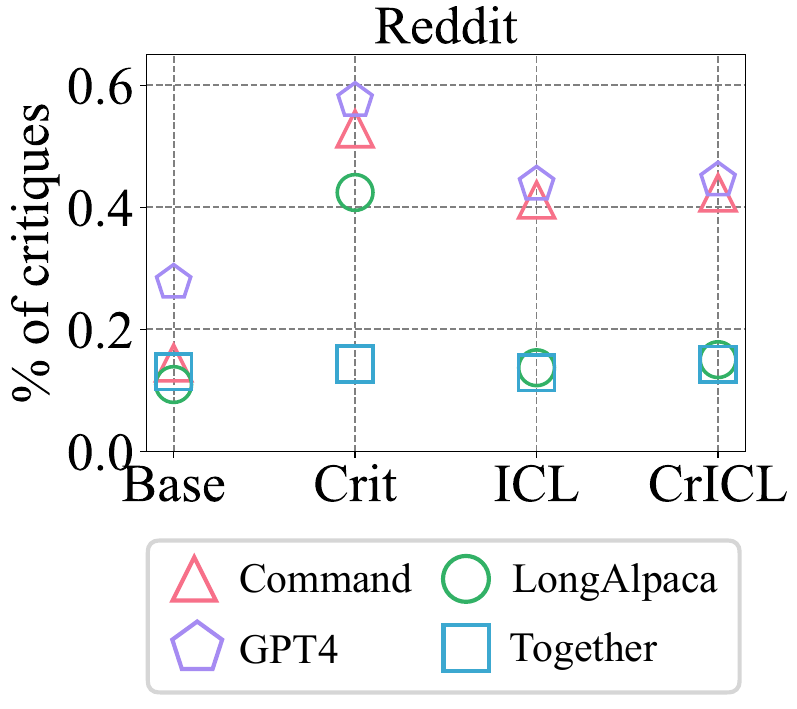}} & \multirow{13}[1]{*}{\includegraphics[scale=0.37]{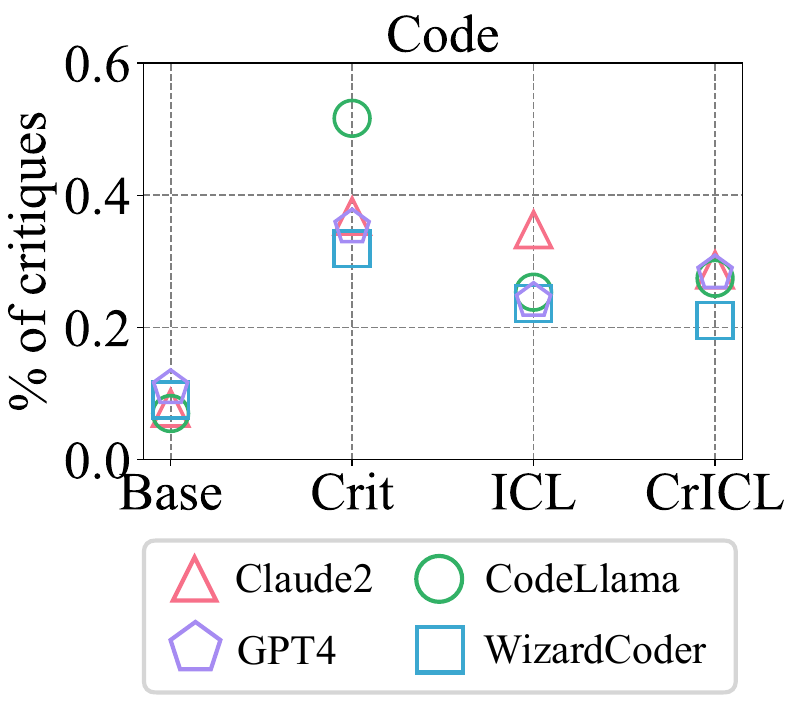}} \\
                                                \\
                                                \\
                                                \\
                                              \\
                                              \\
                                                  \\
                                                   \\
                                                    \\
                                                    \\
                                                    \\
                                                    \\
                                                    \\
                                                 \midrule
                                            & \textbf{Helpfulness} & \\
                                            \midrule
\multirow{13}[1]{*}{\includegraphics[scale=0.37]{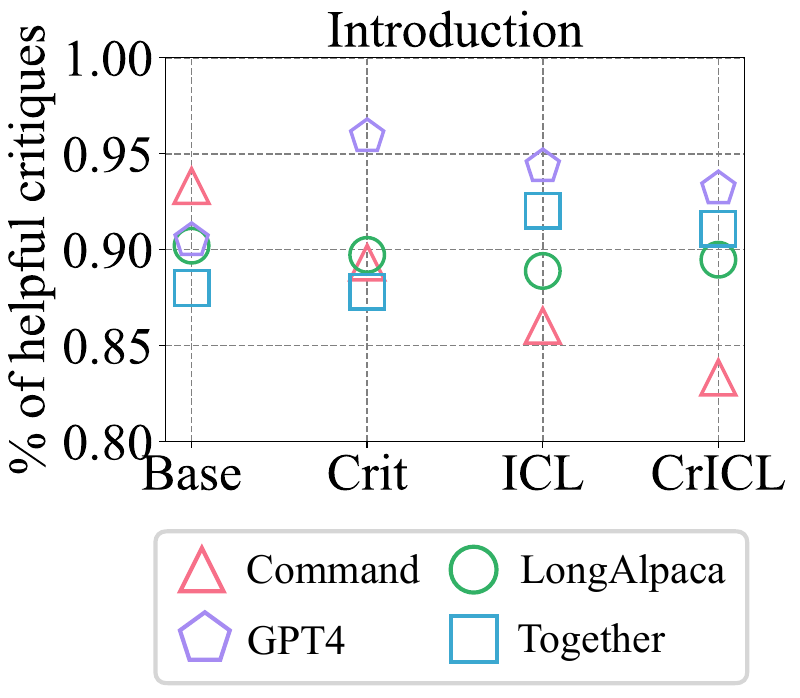}}  & \multirow{13}[1]{*}{\includegraphics[scale=0.37]{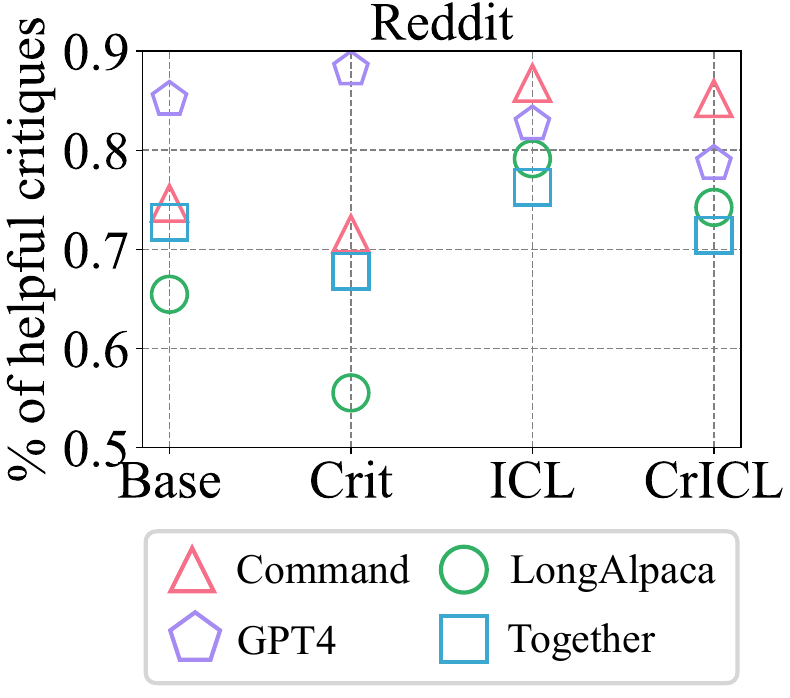}} & \multirow{13}[1]{*}{\includegraphics[scale=0.37]{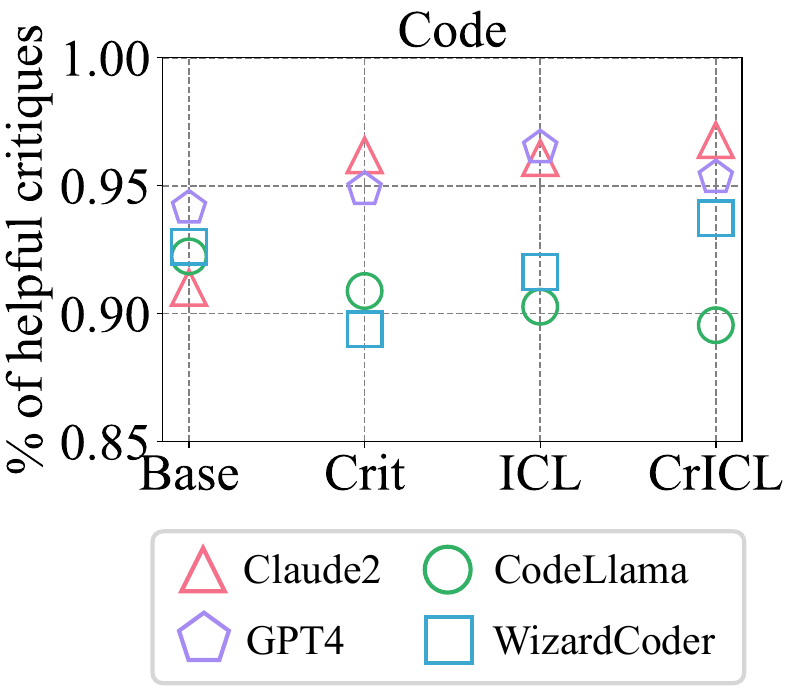}} \\
                                        \\
                                        \\
                                        \\
                                        \\
                                        \\
                                        \\
                                        \\
                                        \\
                                        \\
                                        \\
                                        \\
                                        \\
\bottomrule
\end{tabular}
 \caption{The effect of different strategies on the constructiveness and helpfulness of model-generated feedback texts.}
 \label{tab:lineplot}
\end{table*}
Our study begins by examining LLMs' capabilities in generating valid and contextual feedback texts for given examples. Specifically, we calculate the percentage of feedback texts that are valid and the percentage of valid texts that are contextual. The detailed results are in Appendix Tab.~\ref{tab:validity_context}. Our key findings are as follows: 
(i) Regardless of the strategy used to generate the feedback, \textbf{these LLMs almost always generate valid feedback}, demonstrating their strong instruction-following and in-context learning capability.
(ii) \textbf{The inclusion of demonstrations comes with the risk of distracting models} into writing feedback to the input text of a demonstration, sometimes resulting in a lower overall contextualization of the generated feedback. This is especially true if the demonstrations are long (e.g., introduction task).

\subsection{Constructiveness and Helpfulness}
\label{subsubsec:constructiveness_and_helpfulness}

For feedback texts that passed the validity and contextualization checks, we further scrutinize their constructiveness by counting the percentage of criteria in \S\ref{sec:data_foundataion_for_experiments} they touched upon by providing critiques or suggestions. For feedback texts that passed the constructiveness check as well, we examine their helpfulness by counting the percentage of those critiques or suggestions that are helpful for improvement. Results are in Tab.~\ref{tab:lineplot}.


\paragraph{Constructiveness} (i) For all three tasks across all models, \textbf{the inclusion of criteria generally increases the constructiveness of generated feedback considerably} (on average 25.8\%, at most 44.6\%) , outperforming the base strategy. 
(ii) In most cases, \textbf{adding demonstrations improves constructiveness of generated feedback} compared to the base strategy (on average 10.1\%, at most 27.0\%) . Only in a few cases does the addition of demonstrations lead to a noticeable decrease in constructiveness. This is observed in the introduction task, where the Command and LongAlpaca models produce less constructive feedback with demonstrations, likely struggling with extremely lengthy demonstrations. 
(iii) Interestingly, \textbf{when both demonstrations and criteria are applied, the outcome is mostly comparable to adding demonstrations alone}. This may be attributed to the extended length of demonstrations, which could potentially dilute the impact of the criteria.

\paragraph{Helpfulness} (i) \textbf{The effect of adding criteria on helpfulness is model dependent}. For GPT4, adding criteria always helps to generate more helpful critiques compared to the base strategy. For the other models, on the other hand, in all tasks, adding only criteria typically results in a marginally lower rate of helpful critiques, as compared to the base strategy.
(ii) In most cases, \textbf{adding demonstrations could improve the helpfulness} compared to the base strategy (on average 2.7\%, at most 13.7\%). There are a few exceptions, such as when the base strategy already produces very helpful critiques (over 90\% helpfulness).
(iii) In most cases, \textbf{combining both demonstrations and criteria yields inferior results compared to using demonstrations alone}. This may be attributed to the criteria's lack of specific guidance on crafting helpful critiques and its potential interference with the efficacy of demonstrations when added to the context.

\subsection{Overall Performance}
We also provide the overall performance of each model using different strategies. In particular, we count the total number of criteria the generated feedback texts have addressed by providing helpful critiques or suggestions for the 100 test samples for each task. The results are shown in Tab.~\ref{tab:overall_performance}. We find that (i) \textbf{Providing criteria will always improve the overall performance} regardless of the model and task. (ii) \textbf{Providing demonstrations will generally improve the overall performance} except for the cases where the overly long demonstrations negatively affect the contextualization of feedback too much (in the case of introduction task for most models). (iii) \textbf{Adding both criteria and demonstrations typically does not outperform adding criteria alone}, mostly because models generate fewer critiques/suggestions when demonstrations are provided compared to criteria only (see Appendix Tab.~\ref{tab:overall_performance_critique}). This may be due to the fact that the feedback texts in demonstrations can include positive comments, thus causing the model-generated feedback to be less critical. Therefore, in practice, providing criteria only to LLMs is a reasonably good strategy for feedback generation.

\begin{table}[t]
\centering
\small
  \setlength\tabcolsep{10pt}
  \renewcommand{\arraystretch}{0.8}
\begin{tabular}{lrrrr}
\toprule
\textbf{Intro.}    & Base & Crit & ICL  & CrICL \\
\midrule
Together & 220  & \textbf{324}  & 150  & 133   \\
LAlpaca  & 350  & \textbf{524}  & 200  & 170   \\
Command  & 349  & \textbf{435}  & 117  & 130   \\
GPT4     & 542  & \textbf{983}  & 666  & 656   \\
\midrule
\textbf{Reddit}   & Base & Crit & ICL  & CrICL \\
\midrule
Together & 152  & 200  & 241  & \textbf{245}   \\
LAlpaca  & 165  & \textbf{557}  & 258  & 268   \\
Command  & 265  & \textbf{909}  & 885  & 894   \\
GPT4     & 589  & \textbf{1267} & 904  & 873   \\
\midrule
\textbf{Code}     & Base & Crit & ICL  & CrICL \\
\midrule
CLlama   & 297  & \textbf{1904} & 872  & 848   \\
WizCoder & 389  & \textbf{1095} & 547  & 583   \\
Claude2  & 334  & \textbf{1436} & 1396 & 1051  \\
GPT4     & 481  & \textbf{1494} & 1043 & 1120 \\
\bottomrule
\end{tabular}
\caption{Overall performance of each model using different strategies in terms of the number of criteria the generated feedback texts addressed through providing helpful critiques or suggestions.}
\label{tab:overall_performance}
\end{table}

\section{Further Analysis}
\label{subsection:analysis}
\begin{table}[t]
    \centering
    \small
  \setlength\tabcolsep{0pt}
  \begin{tabular}{cc}
    \toprule
          \multicolumn{2}{c}{\textbf{CodeLlama}}\\
          \midrule
          \multirow{8}[1]{*}{\includegraphics[scale=0.26]{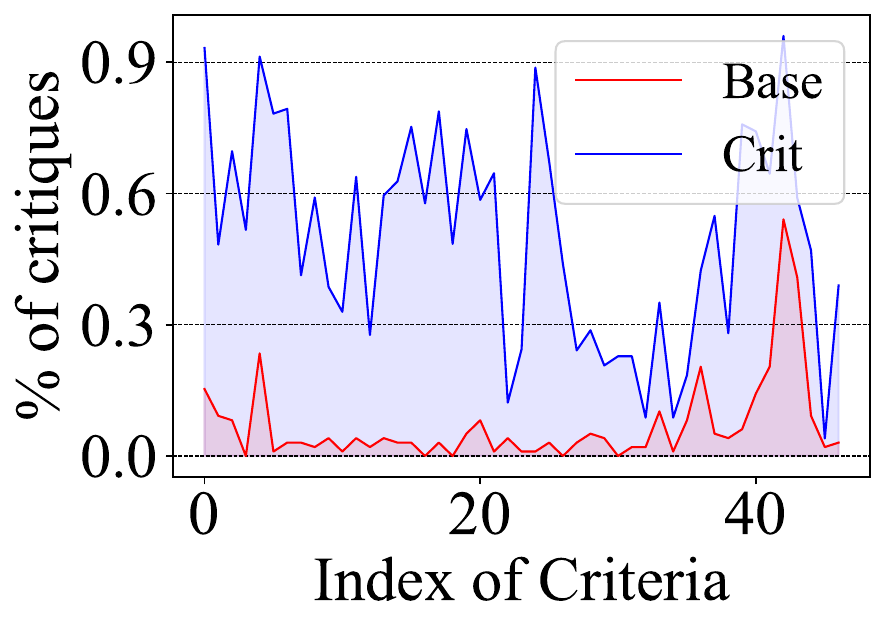}}  & \multirow{8}[1]{*}{\includegraphics[scale=0.26]{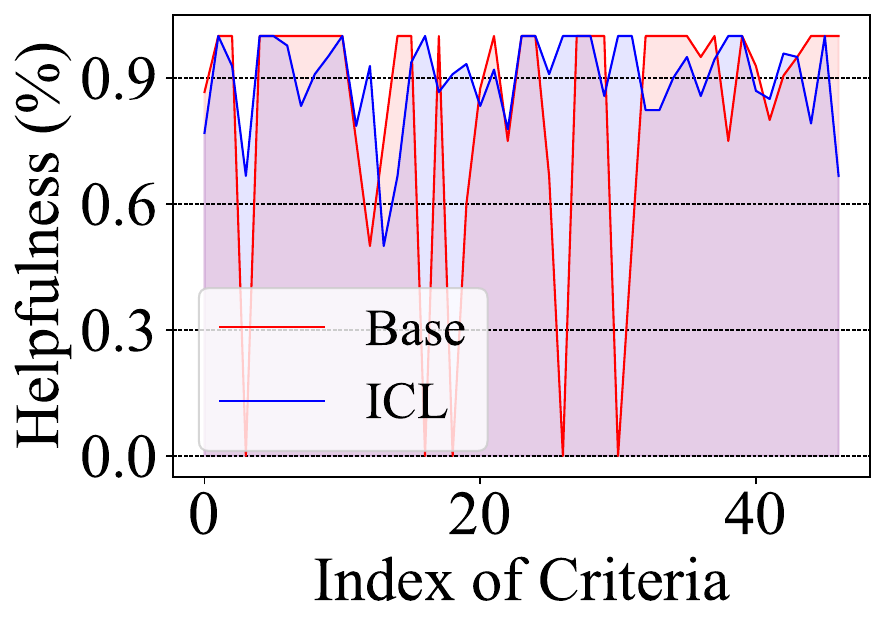}} \\
                                              \\
                                              \\
                                                \\   \\
                                                    \\
                                                    \\
                                                    \\
                                                    \midrule
                \multicolumn{2}{c}{\textbf{WizardCoder}}\\
                \midrule
       \multirow{8}[1]{*}{\includegraphics[scale=0.26]{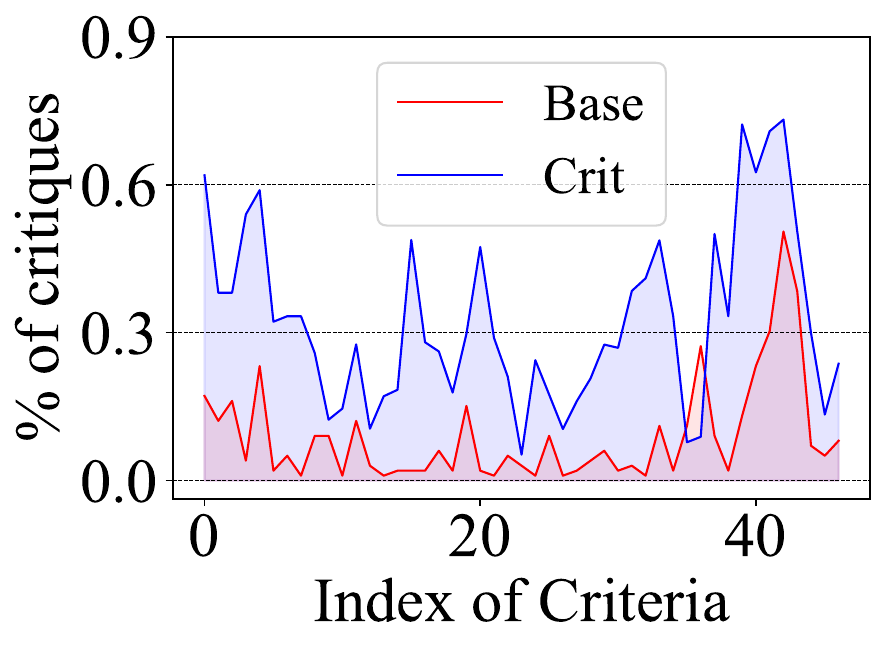}}  & \multirow{8}[1]{*}{\includegraphics[scale=0.26]{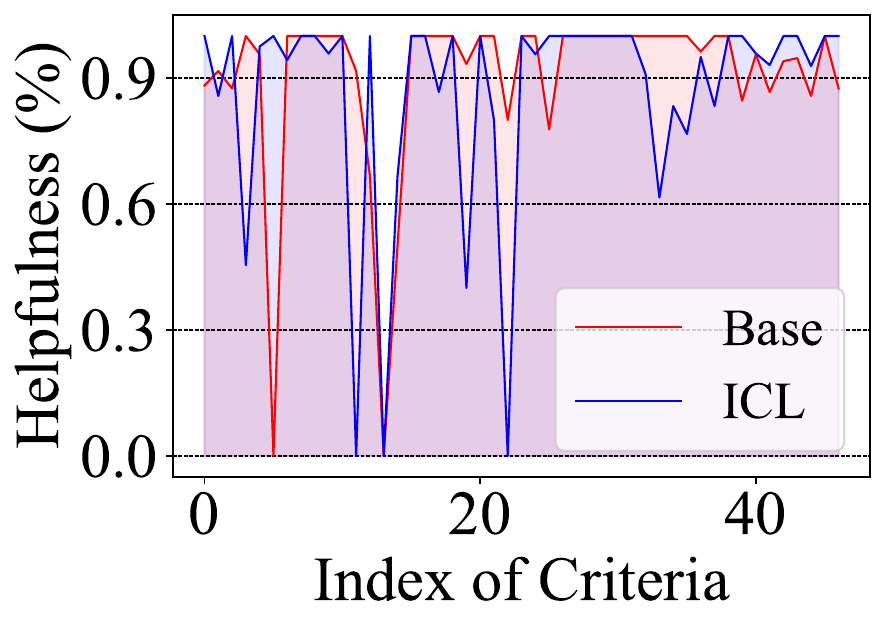}} \\
                                              \\
                                              \\
                                                \\   \\
                                                    \\
                                                    \\
                                                    \\
                                                    \midrule
\multicolumn{2}{c}{\textbf{Claude2}}
       \\
       \midrule
\multirow{8}[1]{*}{\includegraphics[scale=0.26]{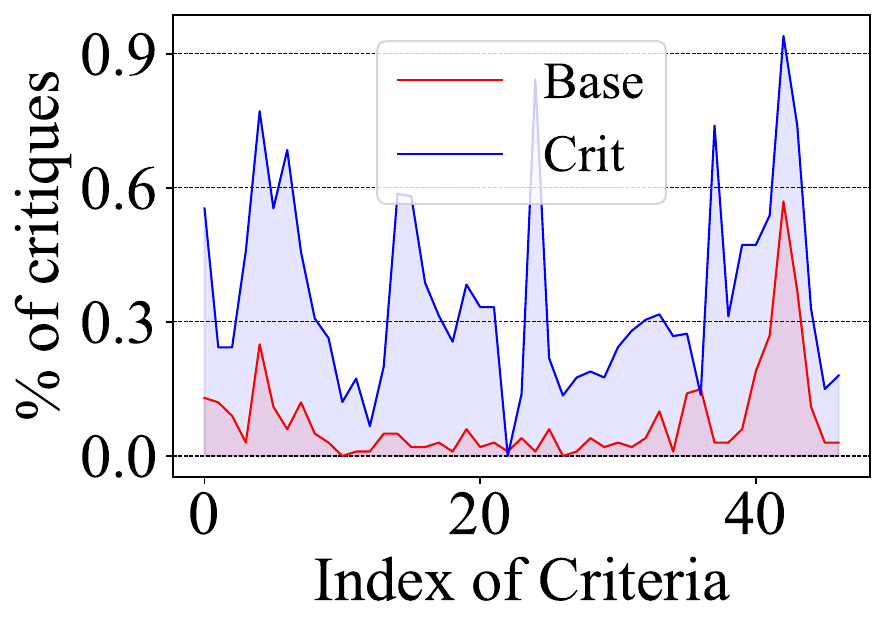}}  & \multirow{8}[1]{*}{\includegraphics[scale=0.26]{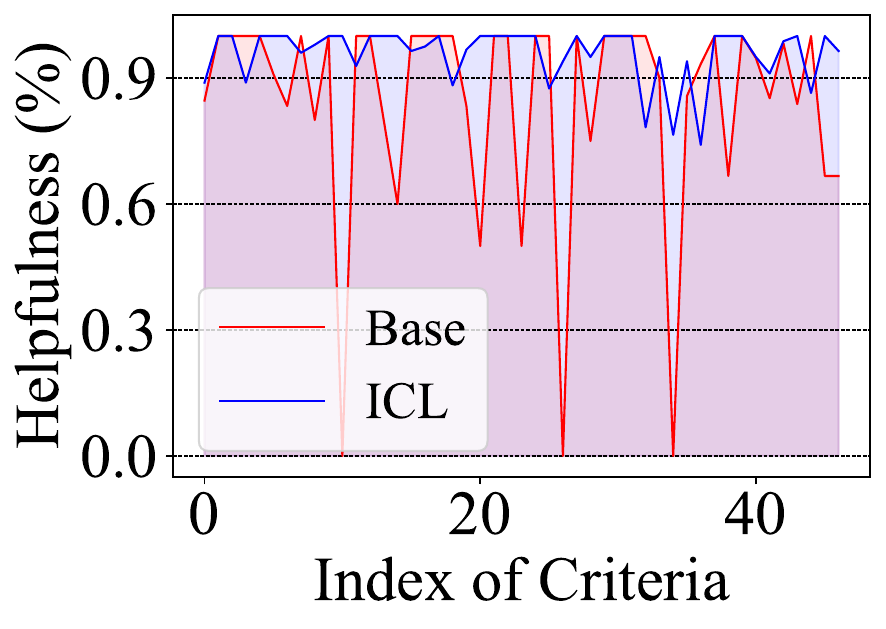}} \\
                                              \\
                                              \\
                                                \\   \\
                                                    \\
                                                    \\
                                                    \\
 \midrule
\multicolumn{2}{c}{\textbf{GPT4}}
       \\
       \midrule
\multirow{8}[1]{*}{\includegraphics[scale=0.26]{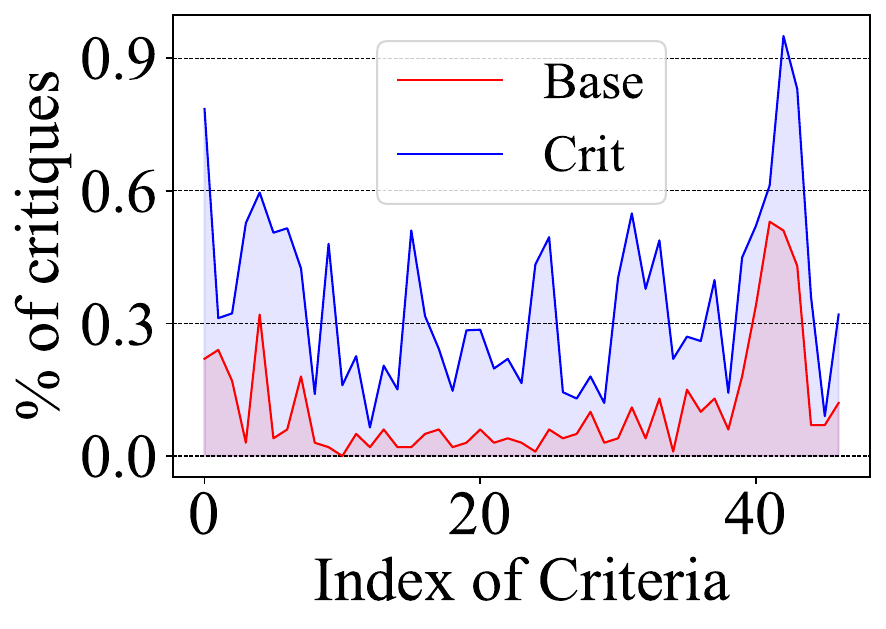}}  & \multirow{8}[1]{*}{\includegraphics[scale=0.26]{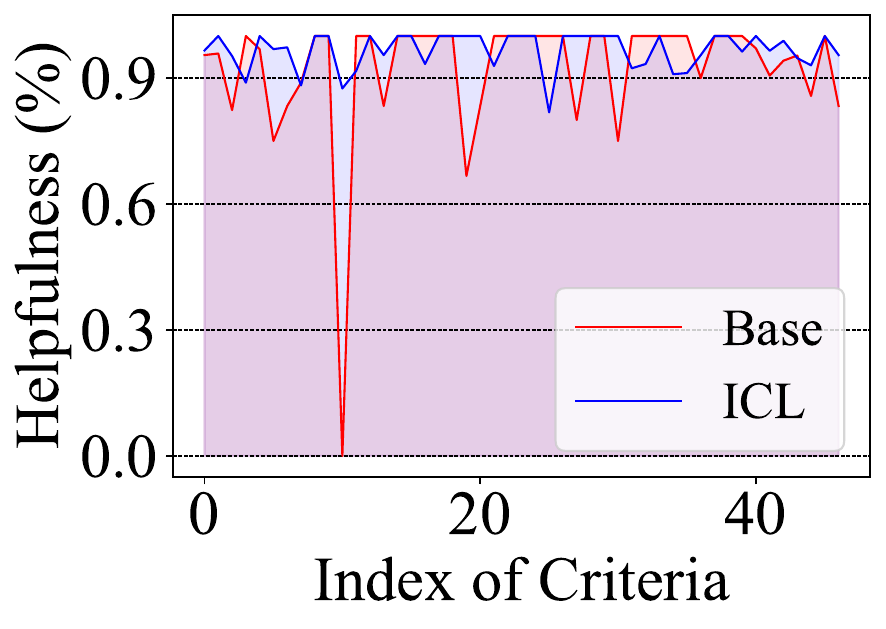}} \\
                                              \\
                                              \\
                                                \\   \\
                                                    \\
                                                    \\
                                                    \\
          \bottomrule
    \end{tabular}
    \caption{Fine-grained analysis of criteria and demonstration effects on the code task. 
    Left: Percentage of critiques out of all critiques that could be generated. Right: Percentage of helpful critiques out of critiques (if there are no critiques, then the number would be 0).
    }
    \label{tab:fine_grained_critique_helpfulness}
\end{table} 

We explore two questions further: (1) How does adding criteria and demonstrations affect the generation of critiques for individual criteria? (2) How providing criteria at different granularities affects the quality of model-generated feedback?


\subsection{Fine-grained Analysis of Criteria and Demonstration Effects}
We selected the code task as our primary testbed, owing to its extensive number of criteria. For each criterion, we analyzed the ratio of generated critiques out of the total critiques that could be generated by the model for 100 test samples. We specifically excluded feedback with validity or contextualization errors. The percentage of generated critiques for each criterion is depicted on the left in Tab.~\ref{tab:fine_grained_critique_helpfulness}. Furthermore, we examined the proportion of helpful critiques within the set of critiques for each criterion, with these percentages presented on the right in Tab.~\ref{tab:fine_grained_critique_helpfulness}. 

We observe that: (i) \textbf{adding criteria enhances the model's ability to generate critiques across various criteria}. Notably, the effectiveness of this enhancement is more pronounced for criteria with which the model is already somewhat familiar, as evidenced by a higher baseline percentage of critiques. (ii) \textbf{Adding demonstrations typically results in a more uniform distribution of helpful critiques across all criteria}, indicating an overall improvement in critique quality. This approach contrasts with the base strategy, which tends to overlook certain criteria.

\subsection{Single vs. Batch Criteria Provision}
To explore the effect of providing criteria at the granularity of an aspect and at the granularity of a single criterion, we chose the Reddit post writing task where each aspect has an average of 4.2 criteria. We only compare the feedback generated by the strategy of offering criteria only. The results are in Tab.~\ref{tab:single_vs_batch}. We see that providing criteria in an aspect's granularity almost always positively affects the validity and contextualization of the generated feedback. However, providing criteria in an aspect's granularity negatively affects the constructiveness of generated feedback, while the impact on helpfulness is model-dependent. In summary, \textbf{providing criteria at the granularity of one aspect rather than on an individual criterion does not significantly affect overall performance, while offering greater efficiency during inference time.}


\begin{table}[t]
\centering
\small
  \setlength\tabcolsep{3pt}
  \renewcommand{\arraystretch}{0.85}
\begin{tabular}{lcccccc}
\toprule
& \multicolumn{3}{c}{Validity}                                                      & \multicolumn{3}{c}{Contextualization}                                             \\
\cmidrule(lr){2-4} \cmidrule(lr){5-7}
& \multicolumn{1}{c}{Single} & \multicolumn{1}{c}{Batch} & \multicolumn{1}{c}{B-S} & \multicolumn{1}{c}{Single} & \multicolumn{1}{c}{Batch} & \multicolumn{1}{c}{B-S} \\
\midrule
Together   & 63.0             & 81.0              & \textbf{+18.0}                       & 60.8                       & 77.8           & \textbf{+17.0}                       \\
LongAlpaca & 78.5                       & 98.5                      & \textbf{+20.0}                       & 74.8                       & 93.0                        & \textbf{+18.2}                     \\
Command    & 95.0                         & 97.8                      & \textbf{+2.8}                      & 93.2                       & 95.0                        & \textbf{+1.8}                      \\
GPT4       & 100.0                        & 100.0                       & 0.0                        & 99.9                       & 100                       & \textbf{+0.1}                      \\
\midrule
& \multicolumn{3}{c}{Constructiveness}                                              & \multicolumn{3}{c}{Helpfulness}                                                   \\
\cmidrule(lr){2-4} \cmidrule(lr){5-7}
& \multicolumn{1}{c}{Single} & \multicolumn{1}{c}{Batch} & \multicolumn{1}{c}{B-S} & \multicolumn{1}{c}{Single} & \multicolumn{1}{c}{Batch} & \multicolumn{1}{c}{B-S} \\
\midrule
Together   & 39.6                       & 14.3                      & -25.3                    & 63.1                       & 67.8                      & \textbf{+4.7}                      \\
LongAlpaca & 48.0                       & 42.4                      & -5.6                     & 62.1                       & 55.5                      & -6.6                     \\
Command    & 69.1                       & 52.9                      & -16.2                    & 77.0                       & 71.6                      & -5.4                     \\
GPT4       & 69.5                       & 57.5                      & -12.0                    & 83.8                       & 88.2                      & \textbf{+4.4}                    \\
\bottomrule
\end{tabular}
\caption{Compare feedback quality when providing criteria in the granularity of a single criterion and in the granularity of an aspect. 
The measurements of ``Validity'', ``Contextualization'', ``Constructiveness'' and ``Helpfulness'' are the same as before.
}
\label{tab:single_vs_batch}
\end{table}

\section{Conclusion}
In this paper, we present a general framework for teaching LLMs to use criteria for feedback generation that takes the guidelines as a starting point, semi-automatically extracts criteria from them, and constructs in-context demonstrations. We then apply these criteria and in-context demonstrations to guide the feedback generation process of LLMs. In order to provide a more comprehensive assessment of the quality of feedback, we propose a layered evaluation methodology to measure the quality of generated feedback from different perspectives. Experiments conducted on three writing tasks and seven different LLMs provide insights on the most effective way of teaching LLMs to use criteria. We also analyzed the effectiveness of applying criteria at different levels of granularity and found that providing criteria at the granularity of an aspect is more effective than providing criteria at the granularity of a single criterion.

\section{Limitations and Future Work}
In our experimental setting, we only consider LLMs that have strong instruction-following capabilities and a large context window. However, there are models that are not as strong at instruction following and in-context learning as the models we have chosen. How to teach those models to use criteria would be an interesting future work. 
In addition, our evaluation is primarily a model-based approach, which may introduce some inaccuracies, although the meta-evaluation shows good correlation with human judgments. Developing better model-based methods to measure feedback quality is also an important direction to explore in the future.

\bibliography{anthology,custom}
\bibliographystyle{acl_natbib}

\appendix

\section{Appendix}
\label{sec:appendix}
\subsection{Prompts For Criteria Extraction}
\label{app:prompt_for_criteria_extraction}
In our paper, we used Claude2 to automatically extract the criteria for each aspect of the guideline (see Tab.~\ref{tab:prompt_for_extracting_criteria} for the prompts used), and then refined and removed duplicates through manual expert review.

\begin{table*}[t]
    \centering
    \footnotesize
  \begin{tabularx}{\textwidth}{@{}X@{}}
    \toprule

\textbf{Prompt for extracting criteria for paper introduction task with Claude2} \\ \midrule
First, I'll share the guidelines for writing scientific papers. Based on the guidelines I gave you, could you please think of a few atomic criteria to use to evaluate a paper? Please only summarize criteria based on the guideline, and do not use your personal knowledge to add additional criteria.\\
\\
\# Guidelines\\
<GUIDELINE>\\
\\
\# Criteria\\
\midrule
\textbf{Prompt for extracting criteria for coding task with Claude2} \\ \midrule
First, I'll share the guidelines for writing Python code. Based on the guideline I gave you, could you please think of a few atomic criteria to use to evaluate a piece of code? Please only summarize criteria based on the guideline, and do not use your personal knowledge to add additional criteria.\\
\\
\# Guidelines\\
<GUIDELINE>\\
\\
\# Criteria\\
\midrule
\textbf{Prompt for extracting criteria for Reddit post task with Claude2} \\ \midrule
First, I'll share the guidelines for creating posts in the WritingPrompts Subreddit community. Based on the guidelines I gave you, could you please think of a few atomic criteria to use to evaluate post submissions? Please only summarize criteria based on the guideline, and do not use your personal knowledge to add additional criteria.\\
\\
\# Guidelines\\
<GUIDELINE>\\
\\
\# Criteria\\
\bottomrule
    \end{tabularx}
    \caption{Prompts for criteria extraction with Claude2. <GUIDELINE> is a placeholder to be filled, which represents a section (aspect) within a guideline.
    }
    \label{tab:prompt_for_extracting_criteria}
\end{table*}

\subsection{Prompts For Demonstration Input Creation}
\label{app:prompt_for_demonstration_input_creation}
In our paper, we use Claude2 to automatically construct demonstration inputs that selectively violate specific criteria in each aspect while adhering to others. Specifically, for a piece of human-written text for a given task, we look at each aspect of the task and use the prompts in Tab.~\ref{tab:prompt_for_ic_input_intro}, Tab.~\ref{tab:prompt_for_ic_input_code} and Tab.~\ref{tab:prompt_for_ic_input_reddit} to iteratively modify the raw text to ultimately obtain demonstration inputs, which are then refined by human expert review.

\begin{table*}[t]
    \centering
    \footnotesize
  \begin{tabularx}{\textwidth}{@{}X@{}}
    \toprule

\textbf{Prompt for constructing \#1 demonstration input for paper introduction task with Claude2} \\ \midrule
Imagine yourself as a professor at a prestigious university who has written hundreds of high-quality academic papers. I'm going to provide you with a set of guidelines for writing high-quality papers. Subsequently, I will share a paper that I've written. Your task is to deliberately modify my paper in a way that contravenes each and every specific criterion listed in the provided guidelines. Additionally, it's crucial that your changes maintain an academic tone and formal language.\\
\\
\# Guidelines for Writing Quality Papers\\
<GUIDELINE>\\
\\
\# Original Paper for Modification\\
<PAPER>\\
\\
\# Specific Criteria to Violate\\
<CRITERIA>\\
\\
\# Modified Paper\\\midrule
\textbf{Prompt for constructing \#2 demonstration input for paper introduction task with Claude2} \\ \midrule
Imagine yourself as a professor at a prestigious university who has written hundreds of high-quality academic papers. I am about to provide you with a set of principles that are essential for writing high-quality papers. Following this, I will present you with a paper that I have authored. I would like you to revise my paper in accordance with each specific criterion delineated in the provided guidelines. If my paper doesn't naturally lend itself to any of the guidelines, please intentionally add elements so it adheres to at least <SAMPLED\_NUMBERINGS> criteria. Ensure that any added text serves a meaningful purpose and contributes to the overall content.\\
\\
\# Guidelines for Writing Quality Papers\\
<GUIDELINE>\\
\\
\# Original Paper for Modification\\
<PAPER>\\
\\
\# Specific Criteria to Adhere To\\
<CRITERIA>\\
\\
\# Modified Paper\\
\bottomrule
    \end{tabularx}
    \caption{Prompts for demonstration input construction with Claude2 for the paper introduction ask. <GUIDELINE> is a placeholder for a section (aspect) of the guideline; <PAPER> is a placeholder for the paper introduction; <CRITERIA> is for all the criteria in an aspect; and <SAMPLED\_NUMBERINGS> is a placeholder for a sampling number of all the criteria in an aspect (e.g., the first, second, and fifth).
    }
    \label{tab:prompt_for_ic_input_intro}
\end{table*}
\begin{table*}[t]
    \centering
    \footnotesize
  \renewcommand{\arraystretch}{0.7}
  \begin{tabularx}{\textwidth}{@{}X@{}}
    \toprule

\textbf{Prompt for constructing \#1 demonstration input for coding task with Claude2} \\ \midrule
Imagine you are a Python engineer with more than 20 years of experience. I'm going to provide you with a set of guidelines for producing high-quality code. Subsequently, I will share a code snippet that I've written. Your task is to deliberately modify my code in a way that contravenes each and every specific criterion listed in the provided guidelines. Importantly, the modifications should serve a meaningful purpose, rather than simply defying the rules without adding any value to the overall functionality.\\
\\
\# Guidelines for Writing Quality Code\\
<GUIDELINE>\\
\\
\# Original Code for Modification\\
\texttt{\textasciigrave\textasciigrave\textasciigrave}python\\
<CODE>\\
\texttt{\textasciigrave\textasciigrave\textasciigrave}\\
\\
\# Specific Criteria to Violate\\
<CRITERIA>\\
\\
\# Modified Code\\
\midrule
\textbf{Prompt for constructing \#2 demonstration input for coding task with Claude2} \\ \midrule
Imagine you are a Python engineer with more than 20 years of experience. I am about to provide you with a set of principles that are essential for writing high-quality code. Following this, I will present you with a code snippet that I have authored. I would like you to revise my code in accordance with each specific criterion delineated in the provided guidelines. If my code doesn't naturally lend itself to any of the guidelines, please intentionally add elements so it adheres to at least <SAMPLED\_NUMBERINGS> criteria. Ensure that any added code serves a meaningful purpose and contributes to the overall functionality.\\
\\
\# Guidelines for Writing Quality Code\\
<GUIDELINE>\\
\\
\# Original Code for Modification\\
\texttt{\textasciigrave\textasciigrave\textasciigrave}python\\
<CODE>\\
\texttt{\textasciigrave\textasciigrave\textasciigrave}\\
\\
\# Specific Criteria to Adhere To\\
<CRITERIA>\\
\\
\# Modified Code\\
\midrule
\textbf{Prompt for constructing \#3 demonstration input for coding task with Claude2} \\ \midrule
Imagine you are a Python engineer with more than 20 years of experience. I'm going to provide you with a set of guidelines for producing high-quality code. Subsequently, I will share a code snippet that I've written. Your task is to deliberately modify my code in a way that contravenes the <SAMPLED\_NUMBERINGS> criteria listed in the provided guidelines. Importantly, the modifications should serve a meaningful purpose, rather than simply defying the rules without adding any value to the overall functionality.\\
\\
\# Guidelines for Writing Quality Code\\
<GUIDELINE>\\
\\
\# Original Code for Modification\\
\texttt{\textasciigrave\textasciigrave\textasciigrave}python\\
<CODE>\\
\texttt{\textasciigrave\textasciigrave\textasciigrave}\\
\\
\# Specific Criteria to Violate\\
<CRITERIA>\\
\\
\# Modified Code\\
\midrule
\textbf{Prompt for constructing \#4 demonstration input for coding task with Claude2} \\ \midrule
Imagine you are a Python engineer with more than 20 years of experience. I am about to provide you with a set of principles that are essential for writing high-quality code. Following this, I will present you with a code snippet that I have authored. I would like you to revise my code in accordance with each specific criterion delineated in the provided guidelines. If my code doesn't naturally lend itself to any of the guidelines, please just return my original code.\\
\\
\# Guidelines for Writing Quality Code\\
<GUIDELINE>\\
\\
\# Original Code for Modification\\
\texttt{\textasciigrave\textasciigrave\textasciigrave}python\\
<CODE>\\
\texttt{\textasciigrave\textasciigrave\textasciigrave}\\
\\
\# Specific Criteria to Adhere To\\
<CRITERIA>\\
\\
\# Modified Code\\
\bottomrule
    \end{tabularx}
    \caption{Prompts for demonstration input construction with Claude2 for the coding task. <GUIDELINE> is a placeholder for a section (aspect) of the guideline; <CODE> is a placeholder for the code snippet; <CRITERIA> is for all the criteria in an aspect; and <SAMPLED\_NUMBERINGS> is a placeholder for a sampling number of all the criteria in an aspect (e.g., the first, second, and fifth).
    }
    \label{tab:prompt_for_ic_input_code}
\end{table*}

\begin{table*}[t]
    \centering
    \footnotesize
  \renewcommand{\arraystretch}{0.7}
  \begin{tabularx}{\textwidth}{@{}X@{}}
    \toprule

\textbf{Prompt for constructing \#1 demonstration input for Reddit post task with Claude2} \\ \midrule
Imagine you're a Community Moderator for the WritingPrompts Subreddit, responsible for ensuring users adhere to the Subreddit's posting rules and guidelines. I'll first outline the guidelines for crafting high-quality posts in the WritingPrompts Subreddit. Then, I'll present a post I've composed. Your challenge is to intentionally revise my post to defy each of the specific criteria highlighted in the guidelines. However, ensure your changes have a purpose and don't merely break the rules without contributing to the content's depth or message.\\
\\
\# Guidelines for Crafting High-Quality Posts\\
<GUIDELINE>\\
\\
\# Original Post for Modification\\
<POST>\\
\\
\# Specific Criteria to Violate\\
<CRITERIA>\\
\\
\# Modified Post\\
\midrule
\textbf{Prompt for constructing \#2 demonstration input for Reddit post task with Claude2} \\ \midrule
Imagine you're a Community Moderator for the WritingPrompts Subreddit, responsible for ensuring users adhere to the Subreddit's posting rules and guidelines. I'll first outline the guidelines for crafting high-quality posts in the WritingPrompts Subreddit. Then, I'll present a post I've composed. I would like you to revise my post in accordance with each specific criterion delineated in the provided guidelines. If my post doesn't naturally lend itself to any of the guidelines, please intentionally add elements so it adheres to at least the <SAMPLED\_NUMBERINGS> criteria. Ensure that any changes serve a meaningful purpose and enhance the post's overall quality.\\
\\
\# Guidelines for Crafting High-Quality Posts\\
<GUIDELINE>\\
\\
\# Original Post for Modification\\
<POST>\\
\\
\# Specific Criteria to Adhere To\\
<CRITERIA>\\
\\
\# Modified Post\\
\midrule
\textbf{Prompt for constructing \#3 demonstration input for Reddit post task with Claude2} \\ \midrule
Imagine you're a Community Moderator for the WritingPrompts Subreddit, responsible for ensuring users adhere to the Subreddit's posting rules and guidelines. I'll first outline the guidelines for crafting high-quality posts in the WritingPrompts Subreddit. Then, I'll present a post I've composed. Your task is to deliberately modify my post in a way that contravenes the <SAMPLED\_NUMBERINGS> criteria listed in the provided guidelines. Importantly, the modifications should serve a meaningful purpose, rather than simply defying the rules without adding any value to the overall content.\\
\\
\# Guidelines for Crafting High-Quality Posts\\
<GUIDELINE>\\
\\
\# Original Post for Modification\\
<POST>\\
\\
\# Specific Criteria to Violate\\
<CRITERIA>\\
\\
\# Modified Post\\
\midrule
\textbf{Prompt for constructing \#4 demonstration input for Reddit post task with Claude2} \\ \midrule
Imagine you're a Community Moderator for the WritingPrompts Subreddit, responsible for ensuring users adhere to the Subreddit's posting rules and guidelines. I'll first outline the guidelines for crafting high-quality posts in the WritingPrompts Subreddit. Then, I'll present a post I've composed. I would like you to revise my post in accordance with each specific criterion delineated in the provided guidelines. If my post doesn't naturally lend itself to any of the guidelines, please just return my original post.\\
\\
\# Guidelines for Crafting High-Quality Posts\\
<GUIDELINE>\\
\\
\# Original Post for Modification\\
<POST>\\
\\
\# Specific Criteria to Adhere To\\
<CRITERIA>\\
\\
\# Modified Post\\
\bottomrule
    \end{tabularx}
    \caption{Prompts for demonstration input construction with Claude2 for the Reddit post task. <GUIDELINE> is a placeholder for a section (aspect) of the guideline; <POST> is a placeholder for the post; <CRITERIA> is for all the criteria in an aspect; and <SAMPLED\_NUMBERINGS> is a placeholder for a sampling number of all the criteria in an aspect (e.g., the first, second, and fifth).
    }
    \label{tab:prompt_for_ic_input_reddit}
\end{table*}

\subsection{Prompts For Demonstration Output Creation}
\label{app:prompt_for_demonstration_output_creation}
In our experiments, we use Claude2 to automatically construct demonstration outputs. In particular, we use prompts in Tab.~\ref{tab:prompt_for_ic_output} to get initial feedback from Claude2, which is then manually refined to address minor issues such as factual accuracy and to enhance the structure and clarity of the feedback.

\begin{table*}[t]
    \centering
    \footnotesize
  \begin{tabularx}{\textwidth}{@{}X@{}}
    \toprule

\textbf{Prompt for constructing demonstration output for paper introduction task with Claude2} \\ \midrule
You are a professor at a prestigious university who has written hundreds of high-quality papers. First, I will provide you with instructions on how to write a good introduction for a research paper. After that, I will present you with my written introduction. Using the guidelines, I would like you to provide as detailed and specific feedback as possible on its strengths and weaknesses, focusing on the specific criteria I've listed.\\
\\
\# Guidelines on how to write a good introduction\\
<GUIDELINE>\\
\\
\# Below is my introduction\\
<INTRODUCTION>\\
\\
\# Criteria to Critique\\
<CRITERIA>\\
\\
\# Your Feedback\\
\midrule
\textbf{Prompt for constructing demonstration output for coding task with Claude2} \\ \midrule
Imagine that you are a Python engineer with over 20 years of experience. I will provide you with a set of guidelines on how to write high-quality code. Then, I will present you with a piece of code that I have written. Using the guidelines, I would like you to provide a detailed and specific critique of the code, focusing on the specific criteria I've listed.\\
\\
\# Guidelines for Writing Quality Code\\
<GUIDELINE>\\
\\
\# My Written Code\\
<CODE>\\
\\
\# Criteria to Critique\\
<CRITERIA>\\
\\
\# Your Feedback\\
\midrule
\textbf{Prompt for constructing demonstration output for Reddit post task with Claude2} \\ \midrule
Imagine you're a Community Moderator for the WritingPrompts Subreddit, responsible for ensuring users adhere to the Subreddit's posting rules and guidelines. I will provide you with a set of guidelines on how to write high-quality posts in the WritingPrompts Subreddit. Then, I will present you with a post that I have written. Using the guidelines, I would like you to provide a detailed and specific critique of the post, focusing on the specific criteria I've listed.\\
\\
\# Guidelines for Crafting High-Quality Posts\\
<GUIDELINE>\\
\\
\# My Written Post\\
<POST>\\
\\
\# Criteria to Critique\\
<CRITERIA>\\
\\
\# Your Feedback\\
\bottomrule
    \end{tabularx}
    \caption{Prompts for demonstration output construction with Claude2. <GUIDELINE> is a placeholder for a section (aspect) of the guideline; <INTRODUCTION>, <CODE>, and <POST> are placeholders for the writing we intend to get feedback for. <CRITERIA> is for all the criteria in an aspect.
    }
    \label{tab:prompt_for_ic_output}
\end{table*}




\subsection{Prompts For Feedback Generation}
\label{app:prompts_for_feedback_generation}
When generating feedback with a given LLM, we use the prompts in Tab.~\ref{tab:prompt_for_generating_feedback_intro}, Tab.~\ref{tab:prompt_for_generating_feedback_code} and Tab.~\ref{tab:prompt_for_generating_feedback_reddit}.

\begin{table*}[t]
    \centering
    \footnotesize
  \renewcommand{\arraystretch}{0.8}
  \begin{tabularx}{\textwidth}{@{}X@{}}
    \toprule

\textbf{Prompt for generating feedback (no criteria, no demonstrations) for paper introduction task} \\ \midrule
You are a professor at a prestigious university who has written hundreds of high-quality papers. I will present you with my written introduction, and I would like you to provide as detailed and specific feedback as possible on its strengths and weaknesses.\\
\\
\# Below is my introduction\\
<INTRODUCTION>\\
\\
\# You should give feedback on my introduction as follows\\
\midrule
\textbf{Prompt for generating feedback (criteria only) for paper introduction task} \\ \midrule
You are a professor at a prestigious university who has written hundreds of high-quality papers. First, I will provide you with guidelines on how to write a good introduction for a research paper. After that, I will present you with my written introduction. Using the guidelines, I would like you to provide as detailed and specific feedback as possible on its strengths and weaknesses, focusing on the specific criteria I've listed.\\
\\
\# Guidelines on how to write a good introduction\\
<GUIDELINE>\\
\\
\# Below is my introduction\\
<INTRODUCTION>\\
\\
\# Criteria to Critique\\
<CRITERIA>\\
\\
\# You should give feedback on my introduction as follows\\
\midrule
\textbf{Prompt for generating feedback (demonstrations only) for paper introduction task} \\ \midrule
You are a professor at a prestigious university who has written hundreds of high-quality papers. First, I will show you two examples on how to judge the quality of an introduction. Then, you should provide feedback on the last introduction.\\
\\
\lbrack Begin Example Introduction\rbrack\\
<EXAMPLE\_INPUT>\\
\lbrack End Example Introduction\rbrack\\
\lbrack Begin Example Feedback\rbrack\\
<EXAMPLE\_OUTPUT>\\
\lbrack End Example Feedback\rbrack\\
-------------------------------------------------------\\
<REMAINING\_DEMONSTRATIONS>\\
-------------------------------------------------------\\
\lbrack Begin Example Introduction\rbrack\\
<USER\_INPUT>\\
\lbrack End Example Introduction\rbrack\\
\lbrack Begin Example Feedback\rbrack\\
\midrule
\textbf{Prompt for generating feedback (with criteria and demonstrations) for paper introduction task} \\ \midrule
You are a professor at a prestigious university who has written hundreds of high-quality papers. First, I will provide you with guidelines on how to write a good introduction for a research paper, along with criteria for evaluating its quality. After that, I will show you two examples of how these guidelines and criteria can be applied to assess the quality of an introduction. Finally, you should provide feedback on the last introduction.\\
\\
\lbrack Begin Guidelines\rbrack\\
<GUIDELINE>\\
\lbrack End Guidelines\rbrack\\
-------------------------------------------------------\\
\lbrack Begin Criteria\rbrack\\
<CRITERIA>\\
\lbrack End Criteria\rbrack\\
-------------------------------------------------------\\
\lbrack Begin Example Introduction\rbrack\\
<EXAMPLE\_INPUT>\\
\lbrack End Example Introduction\rbrack\\
\lbrack Begin Example Feedback\rbrack\\
<EXAMPLE\_OUTPUT>\\
\lbrack End Example Feedback\rbrack\\
-------------------------------------------------------\\
<REMAINING\_DEMONSTRATIONS>\\
-------------------------------------------------------\\
\lbrack Begin Example Introduction\rbrack\\
<USER\_INPUT>\\
\lbrack End Example Introduction\rbrack\\
\lbrack Begin Example Feedback\rbrack\\
\bottomrule
    \end{tabularx}
    \caption{Prompts when generating feedback for the paper introduction task. <GUIDELINE> is a placeholder for a section (aspect) of the guideline; <INTRODUCTION> is a placeholder for the paper introduction; <CRITERIA> is for all the criteria in an aspect; <EXAMPLE\_INPUT> and <EXAMPLE\_OUTPUT> are the input and output of a demonstration. <REMAINING\_DEMONSTRATIONS> is the rest of the demonstrations in the same format as the first one. <USER\_INPUT> is the current text that we want to get feedback on.
    }
    \label{tab:prompt_for_generating_feedback_intro}
\end{table*}

\begin{table*}[t]
    \centering
    \footnotesize
  \renewcommand{\arraystretch}{0.8}
  \begin{tabularx}{\textwidth}{@{}X@{}}
    \toprule

\textbf{Prompt for generating feedback (no criteria, no demonstrations) for coding task} \\ \midrule
Imagine that you are a Python engineer with over 20 years of experience. I will present you with a piece of code that I have written, and I would like you to provide as detailed and specific feedback as possible on its strengths and weaknesses.\\
\\
\# Below is my code\\
<CODE>\\
\\
\# You should give feedback on my code as follows\\
\midrule
\textbf{Prompt for generating feedback (criteria only) for coding task} \\ \midrule
Imagine that you are a Python engineer with over 20 years of experience. I will provide you with a set of guidelines on how to write high-quality code. Then, I will present you with a piece of code that I have written. Using the guidelines, I would like you to provide a detailed and specific critique of the code, focusing on the specific criteria I've listed.\\
\\
\# Guidelines on how to write high-quality code\\
<GUIDELINE>\\
\\
\# Below is my code\\
<CODE>\\
\\
\# Criteria to Critique\\
<CRITERIA>\\
\\
\# You should give feedback on my conclusion as follows\\
\midrule
\textbf{Prompt for generating feedback (demonstrations only) for coding task} \\ \midrule
Imagine that you are a Python engineer with over 20 years of experience. First, I will show you four examples on how to judge the quality of a code snippet. Then, you should provide feedback on the last piece of code. When providing feedback, please adhere to the format used in the earlier demonstration examples.\\
\lbrack Begin Example Code\rbrack\\
<EXAMPLE\_INPUT>\\
\lbrack End Example Code\rbrack\\
\lbrack Begin Example Feedback\rbrack\\
<EXAMPLE\_OUTPUT>\\
\lbrack End Example Feedback\rbrack\\
-------------------------------------------------------\\
<REMAINING\_DEMONSTRATIONS>\\
-------------------------------------------------------\\
\lbrack Begin Example Code\rbrack\\
<USER\_INPUT>\\
\lbrack End Example Code\rbrack\\
\lbrack Begin Example Feedback\rbrack\\
\midrule
\textbf{Prompt for generating feedback (with criteria and demonstrations) for coding task} \\ \midrule
Imagine that you are a Python engineer with over 20 years of experience. First, I will provide you with guidelines on how to write high-quality code, along with criteria for evaluating its quality. After that, I will show you four examples of how these guidelines and criteria can be applied to assess the quality of a code snippet. Finally, you should provide feedback on the last piece of code. When providing feedback, please adhere to the format used in the earlier demonstration examples.\\
\\
\lbrack Begin Guidelines\rbrack\\
<GUIDELINE>\\
\lbrack End Guidelines\rbrack\\
-------------------------------------------------------\\
\lbrack Begin Criteria\rbrack\\
<CRITERIA>\\
\lbrack End Criteria\rbrack\\
-------------------------------------------------------\\
\lbrack Begin Example Code\rbrack\\
<EXAMPLE\_INPUT>\\
\lbrack End Example Code\rbrack\\
\lbrack Begin Example Feedback\rbrack\\
<EXAMPLE\_OUTPUT>\\
\lbrack End Example Feedback\rbrack\\
-------------------------------------------------------\\
<REMAINING\_DEMONSTRATIONS>\\
-------------------------------------------------------\\
\lbrack Begin Example Code\rbrack\\
<USER\_INPUT>\\
\lbrack End Example Code\rbrack\\
\lbrack Begin Example Feedback\rbrack\\

\bottomrule
    \end{tabularx}
    \caption{Prompts when generating feedback for the coding task. <GUIDELINE> is a placeholder for a section (aspect) of the guideline; <CODE> is a placeholder for the code snippet; <CRITERIA> is for all the criteria in an aspect; <EXAMPLE\_INPUT> and <EXAMPLE\_OUTPUT> are the input and output of a demonstration. <REMAINING\_DEMONSTRATIONS> is the rest of the demonstrations in the same format as the first one. <USER\_INPUT> is the current text that we want to get feedback on.
    }
    \label{tab:prompt_for_generating_feedback_code}
\end{table*}

\begin{table*}[t]
    \centering
    \footnotesize
  \renewcommand{\arraystretch}{0.65}
  \begin{tabularx}{\textwidth}{@{}X@{}}
    \toprule

\textbf{Prompt for generating feedback (no criteria, no demonstrations) for Reddit post task} \\ \midrule
Imagine you're a Community Moderator for the WritingPrompts Subreddit, responsible for ensuring users adhere to the Subreddit's posting rules and guidelines. I will present you with a post that I have written, and I would like you to provide as detailed and specific feedback as possible on its strengths and weaknesses.\\
\\
\# Below is my post\\
<POST>\\
\\
\# You should give feedback on my post as follows\\
\midrule
\textbf{Prompt for generating feedback (criteria only) for Reddit post task} \\ \midrule
Imagine you're a Community Moderator for the WritingPrompts Subreddit, responsible for ensuring users adhere to the Subreddit's posting rules and guidelines. I will provide you with a set of guidelines on how to write high-quality posts. Then, I will present you with a post that I have written. Using the guidelines, I would like you to provide a detailed and specific critique of the post, focusing on the specific criteria I've listed.\\
\\
\# Guidelines on how to write high-quality posts\\
<GUIDELINE>\\
\\
\# Below is my post\\
<POST>\\
\\
\# Criteria to Critique\\
<CRITERIA>\\
\\
\# You should give feedback on my post as follows\\
\midrule
\textbf{Prompt for generating feedback (demonstrations only) for Reddit post task} \\ \midrule
Imagine you're a Community Moderator for the WritingPrompts Subreddit, responsible for ensuring users adhere to the Subreddit's posting rules and guidelines. First, I will show you four examples on how to judge the quality of a post. Then, you should provide feedback on the last post. When providing feedback, please adhere to the format used in the earlier demonstration examples.\\
\\
\lbrack Begin Example Post\rbrack\\
<EXAMPLE\_INPUT>\\
\lbrack End Example Post\rbrack\\
\lbrack Begin Example Feedback\rbrack\\
<EXAMPLE\_OUTPUT>\\
\lbrack End Example Feedback\rbrack\\
-------------------------------------------------------\\
<REMAINING\_DEMONSTRATIONS>\\
-------------------------------------------------------\\
\lbrack Begin Example Post\rbrack\\
<USER\_INPUT>\\
\lbrack End Example Post\rbrack\\
\lbrack Begin Example Feedback\rbrack\\
\midrule
\textbf{Prompt for generating feedback (with criteria and demonstrations) for Reddit post task} \\ \midrule
Imagine you're a Community Moderator for the WritingPrompts Subreddit, responsible for ensuring users adhere to the Subreddit's posting rules and guidelines. First, I will provide you with guidelines on how to write high-quality posts, along with criteria for evaluating its quality. After that, I will show you four examples of how these guidelines and criteria can be applied to assess the quality of a post. Finally, you should provide feedback on the last post. When providing feedback, please adhere to the format used in the earlier demonstration examples.\\
\\
\lbrack Begin Guidelines\rbrack\\
<GUIDELINE>\\
\lbrack End Guidelines\rbrack\\
-------------------------------------------------------\\
\lbrack Begin Criteria\rbrack\\
<CRITERIA>\\
\lbrack End Criteria\rbrack\\
-------------------------------------------------------\\
\lbrack Begin Example Post\rbrack\\
<EXAMPLE\_INPUT>\\
\lbrack End Example Post\rbrack\\
\lbrack Begin Example Feedback\rbrack\\
<EXAMPLE\_OUTPUT>\\
\lbrack End Example Feedback\rbrack\\
-------------------------------------------------------\\
<REMAINING\_DEMONSTRATIONS>\\
-------------------------------------------------------\\
\lbrack Begin Example Post\rbrack\\
<USER\_INPUT>\\
\lbrack End Example Post\rbrack\\
\lbrack Begin Example Feedback\rbrack\\
\bottomrule
    \end{tabularx}
    \caption{Prompts when generating feedback for the coding task. <GUIDELINE> is a placeholder for a section (aspect) of the guideline; <POST> is a placeholder for the Reddit post; <CRITERIA> is for all the criteria in an aspect; <EXAMPLE\_INPUT> and <EXAMPLE\_OUTPUT> are the input and output of a demonstration. <REMAINING\_DEMONSTRATIONS> is the rest of the demonstrations in the same format as the first one. <USER\_INPUT> is the current text that we want to get feedback on.
    }
    \label{tab:prompt_for_generating_feedback_reddit}
\end{table*}
\subsection{Prompts For Layered Evaluation}
\label{app:prompts_for_layered_evaluation}


In our study, we used Claude2 to conduct a model-based evaluation of the generated feedback. Our objective is to gauge the helpfulness of the feedback text with respect to each evaluative criterion, specifically focusing on the presence of negative critiques or suggestions pertinent to the criterion that could potentially enhance the quality of the current writing input. To render this evaluation process feasible, we have structured it into the examination of various progressive perspectives, namely validity, contextualization, constructiveness, and helpfulness. The prompts we used for evaluation are detailed in 
Tab.~\ref{tab:prompt_for_evaluating_feedback_validity_intro}\textasciitilde Tab.~\ref{tab:prompt_for_evaluating_feedback_constructiveness_helpfulness_reddit_part2}.

\begin{table*}[t]
    \centering
    \footnotesize
  \renewcommand{\arraystretch}{1}
  \begin{tabularx}{\textwidth}{@{}X@{}}
    \toprule

\textbf{Prompt for evaluating feedback validity for paper introduction task} \\ \midrule
Please evaluate the feedback within the <text> tags. We are looking for feedback that directly addresses parts of the student's introduction section, such as summarizing the content, pointing out strengths, identifying weaknesses, and offering suggestions for improvement. General writing advice without commentary on the specific introduction should not be considered ``specific feedback.''\\
\\
Examples of ``specific'' feedback:\\
1. The introduction briefly overviews cryo-electron microscopy and its ability to determine protein and complex structure.\\
2. The introduction effectively conveys the main points - the problem, approach, and results. It is well-written and understandable.\\
3. The introduction did not identify the research gap the paper aims to fill or justify the paper's importance.\\
4. Your introduction lacks a brief overview that maps out the rest of the paper, which would be useful given the complexity of the topic.\\
\\
Examples of ``not specific'' feedback:\\
1. After getting the reader's attention, add more context and narrow the topic. Only include the most relevant background.\\
2. Directly present the research question with minimal discussion. The rest of the paper will investigate the question.\\
3. The overview should be concise, direct, and in present tense.\\
4. In an empirical research paper, try to lead into the problem on the basis of your discussion of the literature. Think in terms of these questions:\\
\\
<text>\\
<TEXT>\\
</text>\\
\\
Let's evaluate the content within the <text> tags. Describe your thinking step-by-step in the <thinking> tags. Then classify the feedback as ``specific'' or ``not specific'' in the <answer> tags.\\
\bottomrule
    \end{tabularx}
    \caption{Prompt for evaluating feedback validity for the paper introduction task. <TEXT> is a placeholder for the feedback text that we want to evaluate.
    }
    \label{tab:prompt_for_evaluating_feedback_validity_intro}
\end{table*}

\begin{table*}[t]
    \centering
    \footnotesize
  \renewcommand{\arraystretch}{1}
  \begin{tabularx}{\textwidth}{@{}X@{}}
    \toprule

\textbf{Prompt for evaluating feedback contextualization for paper introduction task} \\ \midrule
<scenario>\\
Imagine you are doing a task matching feedback to paper introductions.\\
\\
I will provide:\\
- An introduction enclosed within <introduction> </introduction> tags\\
- Feedback enclosed within <feedback> </feedback> tags\\
\\
Please read both the introduction and feedback carefully. Then make your match determination based on these guidelines:\\
- If the feedback mentions examples, topics or subjects not present in the introduction text, it should be considered ``not match''. However, if the feedback suggests adding content that the current introduction misses, then it should be marked as ``match''\\
- If the feedback seems to refer to other introductions from different papers, it should be considered ``not match''.\\
- If the feedback is too general and you cannot determine if it is specifically about this introduction or not, mark it as ``unsure''.\\
- If the feedback directly comments, critiques or makes suggestions about this particular introduction, it should be considered ``match''.\\
\\
First, describe your step-by-step thinking within <thinking> </thinking> tags.\\
\\
Then provide your final match determination within <answer> </answer> tags using one of the following:\\
- ``match'' if the feedback matches the introduction\\
- ``not match'' if the feedback seems unrelated or about other examples\\
- ``unsure'' if the relationship is unclear\\
\\
Please make your match determination solely based on the current introduction and feedback text provided.\\
</scenario>\\
\\
<introduction>\\
<INTRODUCTION>\\
</introduction>\\
\\
<feedback>\\
<TEXT>\\
</feedback>\\
\bottomrule
    \end{tabularx}
    \caption{Prompt for evaluating feedback contextualization for the paper introduction task. <INTRODUCTION> is the written input text that we want to get feedback on. <TEXT> is a placeholder for the feedback text that we want to evaluate.
    }
    \label{tab:prompt_for_evaluating_feedback_contextualization_intro}
\end{table*}

\begin{table*}[t]
    \centering
    \footnotesize
  \renewcommand{\arraystretch}{1}
  \begin{tabularx}{\textwidth}{@{}X@{}}
    \toprule

\textbf{Prompt for evaluating feedback constructiveness and feedback helpfulness for paper introduction task} \\ \midrule
<scenario>\\
Imagine you are a professor who has written many high-quality research papers. I will provide:\\
- My written introduction enclosed within <introduction> </introduction> tags\\
- Feedback from my advisor on the introduction enclosed within <feedback> </feedback> tags\\
- A criterion the feedback should cover enclosed within <criterion> </criterion> tags\\
\\
Please read the introduction and feedback. Determine if the feedback contains suggestions or negative critique related to the specified criterion. Negative critique means feedback that points out something the introduction does not do well.\\
\\
First, try to extract any suggestions or negative critique relevant to the specified criterion from the feedback into <extraction> </extraction> tags. Your extraction should be as fine-grained as possible, and not include feedback text irrelevant to the current criterion. This may be empty if the feedback has no suggestions or negative critique related to the criterion.\\
\\
Based on whether there is extracted text related to the specified criterion, answer ``yes'' or ``no'' within <negative\_critique\_or\_suggestion> </negative\_critique\_or\_suggestion> tags. If the advisor's feedback contains only positive statements about how the introduction satisfies or adheres to the specified criterion, without any critique or suggestions for improvement, then answer ``no''.\\
\\
If yes: Judge if the extracted suggestions/negative critique provide specific direction for improving the introduction. General suggestions or critique lacking tailored guidance are unhelpful. Also unhelpful is feedback with factual errors about the introduction. Answer ``helpful'' or ``unhelpful'' within <helpfulness> </helpfulness> tags.\\
\\
Critique is unhelpful if it:\\
1) Only reiterates general writing principles without specific guidance. For example, simply stating that the introduction should define the target audience without pointing out issues with how the audience is currently defined would be unhelpful.\\
2) References things not in the introduction or makes inaccurate factual critiques. For example, commenting that the research question is unclear when the research question is explicitly stated would be unhelpful.\\
3) Is vague without clear direction for improvement. For example, saying the motivation should be better explained without elaborating on how to improve the explanation would be unhelpful.\\
4) Does not provide actionable steps for enhancing the introduction. For example, advising the author to overview the literature without specifying what aspects of the literature need more coverage would be unhelpful.\\
\\
If no: Directly answer ``unhelpful'' within <helpfulness> </helpfulness> tags.\\
\\
When you reply, first explain your thought process within <thinking> </thinking> tags. Once you are done thinking, output your final responses within <extraction> </extraction> tags, <negative\_critique\_or\_suggestion> </negative\_critique\_or\_suggestion> tags and <helpfulness> </helpfulness> tags.\\
</scenario>\\
\\
I will show you a few examples of how to make judgments.\\
\\
<example>\\
<introduction>\\
<EXAMPLE\_INTRODUCTION>\\
</introduction>\\
\\
<feedback>\\
<EXAMPLE\_FEEDBACK>\\
</feedback>\\
\\
<criterion>\\
<EXAMPLE\_CRITERION>\\
</criterion>\\
\\
<thinking>\\
<EXAMPLE\_COT1>\\
\\
<extraction>\\
<EXAMPLE\_EXTRACTION>\\
</extraction>\\
\bottomrule
    \end{tabularx}
    \caption{Prompt for evaluating feedback constructiveness and feedback helpfulness for the paper introduction task (part 1). This prompt includes the introduction input of a demonstration <EXAMPLE\_INTRODUCTION>, the demonstration's feedback text <EXAMPLE\_FEEDBACK>, the specific criterion being evaluated <EXAMPLE\_CRITERION>, the CoT reasoning <EXAMPLE\_COT1>, and the extracted relevant text segment <EXAMPLE\_EXTRACTION> from the demonstration.
    }
    \label{tab:prompt_for_evaluating_feedback_constructiveness_helpfulness_intro_part1}
\end{table*}

\begin{table*}[t]
    \centering
    \footnotesize
  \renewcommand{\arraystretch}{1}
  \begin{tabularx}{\textwidth}{@{}X@{}}
    \toprule

\textbf{Prompt for evaluating feedback constructiveness and feedback helpfulness for paper introduction task} \\ \midrule
<EXAMPLE\_COT2>\\
\\
<negative\_critique\_or\_suggestion>\\
<YES\_OR\_NO>\\
</negative\_critique\_or\_suggestion>\\
\\
<EXAMPLE\_COT3>\\
\\
<helpfulness>\\
<HELPFUL\_OR\_UNHELPFUL>\\
</helpfulness>\\
</thinking>\\
</example>\\
-------------------------------------------------------\\
<REMAINING\_DEMONSTRATIONS>\\
-------------------------------------------------------\\
Below is my actual introduction, my received feedback, and the criterion. Now it's your turn to make judgments.\\
<introduction>\\
<INTRODUCTION>\\
</introduction>\\
\\
<feedback>\\
<FEEDBACK>\\
</feedback>\\
\\
<criterion>\\
<CRITERION>\\
</criterion>\\
\bottomrule
    \end{tabularx}
    \caption{Prompt for evaluating feedback constructiveness and feedback helpfulness for the paper introduction task (part 2). This part covers the CoT reasoning (<EXAMPLE\_COT2>, <EXAMPLE\_COT3>), decisions on the presence of constructive or critical suggestions (<YES\_OR\_NO>), and the determination of their helpfulness (<HELPFUL\_OR\_UNHELPFUL>). It also includes the remaining demonstrations (<REMAINING\_DEMONSTRATIONS>) formatted similarly to the initial demonstration, the current introduction input (<INTRODUCTION>), the feedback text (<FEEDBACK>) under review, and the specific evaluation criterion (<CRITERION>).
    }
    \label{tab:prompt_for_evaluating_feedback_constructiveness_helpfulness_intro_part2}
\end{table*}

\begin{table*}[t]
    \centering
    \footnotesize
  \renewcommand{\arraystretch}{1}
  \begin{tabularx}{\textwidth}{@{}X@{}}
    \toprule

\textbf{Prompt for evaluating feedback validity for coding task} \\ \midrule
Please evaluate the feedback within the <text> tags. We are looking for feedback that directly addresses parts of the engineer's code.\\
\\
Feedback is ``not specific'' if any of the following applies:\\
- It only provides modified code without critiquing the original first.\\
- It gives general coding advice without specifics for the code provided.\\
- It does not relate the advice back to the original code.\\
\\
Feedback is ``specific'' if any of the following applies:\\
- It summarizes the content.\\
- It points out strengths or identifies weaknesses.\\
- It identifies where the code diverges from best practices and suggestions to improve it.\\
- It explains why the guidelines do not apply to the current code.\\
- It contains both specific and non-specific elements, it should be considered overall as ``specific.''\\
\\
Examples of ``specific'' feedback:\\
1. The `favoriteColor' variable is not used anywhere in the code, so it can be removed.\\
2. The code does not make use of built-in exception classes for argument validation, but instead uses `assert' statements. This is not recommended as `assert' is used to ensure internal correctness, not to enforce correct usage nor to indicate that some unexpected event occurred.\\
3. The code is using simple comprehensions and generator expressions, so it is okay to use them for simple cases.\\
4. The `BlackBelt' class uses the `list\_commands' and `get\_command' methods as class methods, which is appropriate.\\
\\
Examples of ``not specific'' feedback:\\
1. Here is a modified version of the code with some fixes: [provides modified code]. The modified code uses more descriptive name and follows Python style guidelines.\\
2. Use `classmethod' only when writing a named constructor, or a class-specific routine that modifies necessary global state such as a process-wide cache.\\
3. You should never compare a boolean variable to `False' using `=='.\\
4. I'm happy to help, but I cannot give feedback on your code. What else can I help you with?\\
\\
<text>\\
<TEXT>\\
</text>\\
\\
Let's evaluate the content within the <text> tags. Describe your thinking step-by-step in the <thinking> tags. Then classify the feedback as ``specific'' or ``not specific'' in the <answer></answer> tags.\\
\bottomrule
    \end{tabularx}
    \caption{Prompt for evaluating feedback validity for the coding task. <TEXT> is a placeholder for the feedback text that we want to evaluate.
    }
    \label{tab:prompt_for_evaluating_feedback_validity_code}
\end{table*}

\begin{table*}[t]
    \centering
    \footnotesize
  \renewcommand{\arraystretch}{1}
  \begin{tabularx}{\textwidth}{@{}X@{}}
    \toprule

\textbf{Prompt for evaluating feedback contextualization for coding task} \\ \midrule
<scenario>\\
Imagine you are doing a task matching feedback to code.\\
\\
I will provide:\\
- A code snippet enclosed within <code> </code> tags\\
- Feedback enclosed within <feedback> </feedback> tags\\
\\
Please read both the code and feedback carefully. Then make your match determination based on these guidelines:\\
- If the feedback references functions, variables, classes, code examples etc. **not present** in the provided code snippet, mark it as ``not match''. However, if the feedback suggests adding content that the current code misses, then it should be marked as ``match''\\
- If the feedback refers to other code examples or code snippets (e.g., the first example ...), it should be considered ``not match''.\\
- If the feedback is too general and you cannot determine if it is specifically about this code or not, mark it as ``unsure''.\\
- If the feedback directly critiques or makes suggestions about this particular code, it should be considered ``match''.\\
- If the feedback comments on the code (e.g. summarizes what it does), mark it as ``match''.\\
- If the feedback points out that the code **does not** do something, then mark it as ``match''\\
- If the feedback is a mixture of ``match'' and ``not match'' content, then mark it as ``match''\\
\\
First, describe your step-by-step thinking within <thinking> </thinking> tags.\\
\\
Then provide your final match determination within <answer> </answer> tags using one of the following:\\
\\
- ``match'' if the feedback matches the code\\
- ``not match'' if the feedback seems unrelated or about other code examples\\
- ``unsure'' if the relationship is unclear\\
\\
Please make your match determination solely based on the current code and feedback text provided.\\
</scenario>\\
\\
<code>\\
<CODE>\\
</code>\\
\\
<feedback>\\
<TEXT>\\
</feedback>\\
\bottomrule
    \end{tabularx}
    \caption{Prompt for evaluating feedback contextualization for the coding task. <CODE> is the written code that we want to get feedback on. <TEXT> is a placeholder for the feedback text that we want to evaluate.
    }
    \label{tab:prompt_for_evaluating_feedback_contextualization_code}
\end{table*}

\begin{table*}[t]
    \centering
    \footnotesize
  \renewcommand{\arraystretch}{1}
  \begin{tabularx}{\textwidth}{@{}X@{}}
    \toprule

\textbf{Prompt for evaluating feedback constructiveness and feedback helpfulness for coding task} \\ \midrule
<scenario>\\
Imagine you are a Python engineer with over 20 years of experience. I will provide:\\
\\
- My written Python code enclosed within <code> </code> tags\\
- Feedback from my manager on the code enclosed within <feedback> </feedback> tags\\
- A criterion the feedback should cover enclosed within <criterion> </criterion> tags\\
\\
Please read the code and feedback. Determine if the feedback contains suggestions or negative critique related to the specified criterion. Negative critique means feedback that points out something the code does not do well.\\
\\
First, try to extract any suggestions or negative critique relevant to the criterion from the feedback into <extraction> </extraction> tags. Your extraction should be as fine-grained as possible, and not include feedback text irrelevant to the current criterion. This may be empty if the feedback has no suggestions or negative critique related to the criterion.\\
\\
Based on whether there is extracted text related to the specified criterion, answer ``yes'' or ``no'' within <negative\_critique\_or\_suggestion> </negative\_critique\_or\_suggestion> tags. If the manager's feedback contains only positive statements about how the code satisfies or adheres to the specified criterion, without any critique or suggestions for improvement, then answer ``no''.\\
\\
If yes: Judge if the extracted suggestions/negative critique provide specific direction for improving the code. General suggestions or critique lacking tailored guidance are unhelpful. Also unhelpful is feedback with factual errors about the code. Answer ``helpful'' or ``unhelpful'' within <helpfulness> </helpfulness> tags.\\
\\
Critique/suggestion is unhelpful if it\\
1) Only reiterates general writing principles without specific guidance. For example, mentioning the importance of naming conventions without specifying which part of the code violates these conventions.\\
2) Refer to code snippets that aren't included in the provided <code> tags, or make incorrect assumptions about the code's functionality.\\
3) Incorrectly state that the code is missing certain elements or functionalities when, in reality, these are already present in the code.\\
\\
If no: Directly answer ``unhelpful'' within <helpfulness> </helpfulness> tags.\\
\\
When you reply, first explain your thought process within <thinking> </thinking> tags. Once you are done thinking, output your final responses within <extraction> </extraction> tags, <negative\_critique\_or\_suggestion> </negative\_critique\_or\_suggestion> tags and <helpfulness> </helpfulness> tags.\\
</scenario>\\
\\
I will show you a few examples of how to make judgments.\\
\\
<example>\\
<EXAMPLE\_CODE>\\
</code>\\
\\
<feedback>\\
<EXAMPLE\_FEEDBACK>\\
</feedback>\\
\\
<criterion>\\
<EXAMPLE\_CRITERION>\\
</criterion>\\
\\
<thinking>\\
<EXAMPLE\_COT1>\\
\\
<extraction>\\
<EXAMPLE\_EXTRACTION>\\
</extraction>\\
\bottomrule
    \end{tabularx}
    \caption{Prompt for evaluating feedback constructiveness and feedback helpfulness for the coding task (part 1). This prompt includes the code input of a demonstration <EXAMPLE\_CODE>, the demonstration's feedback text <EXAMPLE\_FEEDBACK>, the specific criterion being evaluated <EXAMPLE\_CRITERION>, the CoT reasoning <EXAMPLE\_COT1>, and the extracted relevant text segment <EXAMPLE\_EXTRACTION> from the demonstration.
    }
    \label{tab:prompt_for_evaluating_feedback_constructiveness_helpfulness_code_part1}
\end{table*}

\begin{table*}[t]
    \centering
    \footnotesize
  \renewcommand{\arraystretch}{1}
  \begin{tabularx}{\textwidth}{@{}X@{}}
    \toprule

\textbf{Prompt for evaluating feedback constructiveness and feedback helpfulness for coding task} \\ \midrule
<EXAMPLE\_COT2>\\
\\
<negative\_critique\_or\_suggestion>\\
<YES\_OR\_NO>\\
</negative\_critique\_or\_suggestion>\\
\\
<EXAMPLE\_COT3>\\
\\
<helpfulness>\\
<HELPFUL\_OR\_UNHELPFUL>\\
</helpfulness>\\
</thinking>\\
</example>\\
-------------------------------------------------------\\
<REMAINING\_DEMONSTRATIONS>\\
-------------------------------------------------------\\
Below is my actual code, my received feedback and the criterion.\\
<code>\\
<CODE>\\
</code>\\
\\
<feedback>\\
<FEEDBACK>\\
</feedback>\\
\\
<criterion>\\
<CRITERION>\\
</criterion>\\
\bottomrule
    \end{tabularx}
    \caption{Prompt for evaluating feedback constructiveness and feedback helpfulness for the coding task (part 2). This part covers the CoT reasoning (<EXAMPLE\_COT2>, <EXAMPLE\_COT3>), decisions on the presence of constructive or critical suggestions (<YES\_OR\_NO>), and the determination of their helpfulness (<HELPFUL\_OR\_UNHELPFUL>). It also includes the remaining demonstrations (<REMAINING\_DEMONSTRATIONS>) formatted similarly to the initial demonstration, the current code input (<CODE>), the feedback text (<FEEDBACK>) under review, and the specific evaluation criterion (<CRITERION>).
    }
    \label{tab:prompt_for_evaluating_feedback_constructiveness_helpfulness_code_part2}
\end{table*}

\begin{table*}[t]
    \centering
    \footnotesize
  \renewcommand{\arraystretch}{1}
  \begin{tabularx}{\textwidth}{@{}X@{}}
    \toprule

\textbf{Prompt for evaluating feedback validity for Reddit post task} \\ \midrule
Please evaluate the feedback within the <text> tags.\\
\\
Feedback is ``not specific'' if any of the following applies:\\
- Only provides general advice about writing good prompts\\
- Does not directly reference, evaluate or critique this example post\\
- Discusses prompt writing guidelines without applying them to this post\\
\\
Feedback is ``specific'' if any of the following applies:\\
- Summarizes this post's content in present tense\\
- States that the prompt follows or violates a guideline.\\
- Points out strengths or weaknesses in this particular post\\
- Offers suggestions to improve this specific post\\
- It explains why the guidelines do not apply to the current post.\\
- It contains both specific and non-specific elements, it should be considered overall as ``specific.''\\
\\
Examples of ``specific'' feedback:\\
1. The post does not contain any external links or references to other subreddits or crowdfunding platforms.\\
2. Overall, the title is well-crafted and suitable for inspiring a wide range of narratives and poems.\\
3. The post directly involves sharing confidential information found in the supposed journals of Elon Musk. This could be interpreted as encouraging the dissemination of personal or confidential information, which is a violation of the guidelines on personal information protection.\\
4. The prompt directly involves real-world Reddit users, which is against the guidelines.\\
\\
Examples of ``not specific'' feedback:\\
1. Crafting open-ended prompts is essential for fostering creativity. Such prompts should be designed to be interpreted in a multitude of ways, allowing writers the freedom to explore diverse narratives and themes.\\
2. Prompts should center around fostering creativity and writing, not competition or financial gain.\\
3. Prompts must not encourage, contain, or lead to illegal content or activities.\\
4. [emojis]\\
\\
\\
<text>\\
<TEXT>\\
</text>\\
\\
Let's evaluate the content within the <text> tags. Does this provide specific feedback on the qualities and content of the Reddit post? Explain your reasoning in the <thinking> tags. Then give a ``specific'' or ``not specific'' answer in the <answer> tags. Do not give other answers.\\
\bottomrule
    \end{tabularx}
    \caption{Prompt for evaluating feedback validity for the Reddit post task. <TEXT> is a placeholder for the feedback text that we want to evaluate.
    }
    \label{tab:prompt_for_evaluating_feedback_validity_reddit}
\end{table*}

\begin{table*}[t]
    \centering
    \footnotesize
  \renewcommand{\arraystretch}{1}
  \begin{tabularx}{\textwidth}{@{}X@{}}
    \toprule

\textbf{Prompt for evaluating feedback contextualization for Reddit post task} \\ \midrule
<scenario>\\
Imagine you are doing a task matching feedback to a Reddit post.\\
\\
I will provide:\\
- A Reddit post enclosed within <post> </post> tags\\
- Feedback enclosed within <feedback> </feedback> tags\\
\\
Please read both the post and feedback carefully. Then make your match determination based on these guidelines:\\
- If the feedback references content **not present** in the provided Reddit post, mark it as ``not match''. However, if the feedback suggests adding content that the current post misses, then it should be marked as ``match''\\
- If the feedback refers to other post examples (e.g., the first example ...), it should be considered ``not match''.\\
- If the feedback is general and you cannot determine if it is specifically about this post or not, mark it as ``unsure''.\\
- If the feedback directly critiques or makes suggestions about this particular post, it should be considered ``match''.\\
- If the feedback comments on the post (e.g. summarizes what it does), mark it as ``match''.\\
- If the feedback points out that the post **does not** do something, then mark it as ``match''\\
- If the feedback states that the prompt follows or violates a guideline without giving specific evidence, mark it as ``match''\\
- If the feedback is a mixture of ``match'' and ``not match'' content, then mark it as ``match''\\
- If you cannot find evidence that shows ``not match'' or ``match'', then mark it ``unsure''.\\
\\
First, describe your step-by-step thinking within <thinking> </thinking> tags. Then provide your final match determination within <answer> </answer> tags using one of the following:\\
- ``match'' if the feedback matches the post\\
- ``not match'' if the feedback seems unrelated or about other post examples\\
- ``unsure'' if the relationship is unclear\\
\\
Please make your match determination solely based on the current post and feedback text provided.\\
</scenario>\\
\\
<post>\\
<POST>\\
</post>\\
\\
<feedback>\\
<TEXT>\\
</feedback>\\
\bottomrule
    \end{tabularx}
    \caption{Prompt for evaluating feedback contextualization for the Reddit post task. <POST> is the written post that we want to get feedback on. <TEXT> is a placeholder for the feedback text that we want to evaluate.
    }
    \label{tab:prompt_for_evaluating_feedback_contextualization_reddit}
\end{table*}

\begin{table*}[t]
    \centering
    \footnotesize
  \renewcommand{\arraystretch}{1}
  \begin{tabularx}{\textwidth}{@{}X@{}}
    \toprule

\textbf{Prompt for evaluating feedback constructiveness and feedback helpfulness for Reddit post task} \\ \midrule
<scenario>\\
Imagine you're a Community Moderator for the WritingPrompts Subreddit, responsible for ensuring users adhere to the Subreddit's posting rules and guidelines. I will provide:\\
\\
- My written post enclosed within <post> </post> tags\\
- Feedback from my friend on the post enclosed within <feedback> </feedback> tags\\
- A criterion the feedback should cover enclosed within <criterion> </criterion> tags\\
\\
Please read the post and feedback. Determine if the feedback contains suggestions or negative critique related to the specified criterion. Negative critique means feedback that points out something the post does not do well.\\
\\
First, try to extract any suggestions or negative critique relevant to the specified criterion from the feedback into <extraction> </extraction> tags. You extraction should be as fine-grained as possible, and not include feedback text irrelevant to the current criterion. This may be empty if the feedback has no suggestions or negative critique related to the criterion.\\
\\
Based on whether there is extracted text related to the specified criterion, answer ``yes'' or ``no'' within <negative\_critique\_or\_suggestion> </negative\_critique\_or\_suggestion> tags. If the friend's feedback contains only positive statements about how the post satisfies or adheres to the specified criterion, without any critique or suggestions for improvement, then answer ``no''.\\
\\
If yes: Judge if the extracted suggestions/negative critique provide specific direction for improving the post and whether those direction can actually make the post more aligned with the criterion. Answer ``helpful'' or ``unhelpful'' within <helpfulness> </helpfulness> tags.\\
\\
Critique or suggestion is unhelpful if it:\\
1) Only reiterates general writing principles without specific guidance. For example, simply stating that the prompt should be more creative without providing specific suggestions on how to make it more creative.\\
2) References things not in the post or makes inaccurate factual critiques. For example, commenting that the post contains offensive language when it does not.\\
3) Is vague without clear direction for improvement. For example, advising the author to make the title more interesting without elaborating on how to make it more interesting.\\
4) Does not provide actionable steps for enhancement. For example, telling the author to expand on the backstory without specifying which parts of the backstory need more detail.\\
\\
If no: Directly answer ``unhelpful'' within <helpfulness> </helpfulness> tags.\\
\\
When you reply, first explain your thought process within <thinking> </thinking> tags. Once you are done thinking, output your final responses within <extraction> </extraction> tags, <negative\_critique\_or\_suggestion> </negative\_critique\_or\_suggestion> tags and <helpfulness> </helpfulness> tags.\\
</scenario>\\
\\
Below are some examples\\
<example>\\
<post>\\
<EXAMPLE\_POST>\\
</post>\\
\\
<feedback>\\
<EXAMPLE\_FEEDBACK>\\
</feedback>\\
\\
<criterion>\\
<EXAMPLE\_CRITERION>\\
</criterion>\\
\\
<thinking>\\
<EXAMPLE\_COT1>\\
\\
<extraction>\\
<EXAMPLE\_EXTRACTION>\\
</extraction>\\
\bottomrule
    \end{tabularx}
    \caption{Prompt for evaluating feedback constructiveness and feedback helpfulness for the Reddit post task (part 1). This prompt includes the post input of a demonstration <EXAMPLE\_POST>, the demonstration's feedback text <EXAMPLE\_FEEDBACK>, the specific criterion being evaluated <EXAMPLE\_CRITERION>, the CoT reasoning <EXAMPLE\_COT1>, and the extracted relevant text segment <EXAMPLE\_EXTRACTION> from the demonstration.
    }
    \label{tab:prompt_for_evaluating_feedback_constructiveness_helpfulness_reddit_part1}
\end{table*}

\begin{table*}[t]
    \centering
    \footnotesize
  \renewcommand{\arraystretch}{1}
  \begin{tabularx}{\textwidth}{@{}X@{}}
    \toprule

\textbf{Prompt for evaluating feedback constructiveness and feedback helpfulness for Reddit post task} \\ \midrule
<EXAMPLE\_COT2>\\
\\
<negative\_critique\_or\_suggestion>\\
<YES\_OR\_NO>\\
</negative\_critique\_or\_suggestion>\\
\\
<EXAMPLE\_COT3>\\
\\
<helpfulness>\\
<HELPFUL\_OR\_UNHELPFUL>\\
</helpfulness>\\
</thinking>\\
</example>\\
-------------------------------------------------------\\
<REMAINING\_DEMONSTRATIONS>\\
-------------------------------------------------------\\
Below is my actual post, my received feedback and the criterion.\\
<post>\\
<POST>\\
</post>\\
\\
<feedback>\\
<FEEDBACK>\\
</feedback>\\
\\
<criterion>\\
<CRITERION>\\
</criterion>\\
\bottomrule
    \end{tabularx}
    \caption{Prompt for evaluating feedback constructiveness and feedback helpfulness for the Reddit post task (part 2). This part covers the CoT reasoning (<EXAMPLE\_COT2>, <EXAMPLE\_COT3>), decisions on the presence of constructive or critical suggestions (<YES\_OR\_NO>), and the determination of their helpfulness (<HELPFUL\_OR\_UNHELPFUL>). It also includes the remaining demonstrations (<REMAINING\_DEMONSTRATIONS>) formatted similarly to the initial demonstration, the current post input (<POST>), the feedback text (<FEEDBACK>) under review, and the specific evaluation criterion (<CRITERION>).
    }
    \label{tab:prompt_for_evaluating_feedback_constructiveness_helpfulness_reddit_part2}
\end{table*}

\subsection{Meta-Evaluation for Our Evaluation Methodology}
\label{app:meta-eval}
While our model-based evaluation demonstrates high accuracy in alignment with human judgments, we have also undertaken a detailed analysis of human label distribution within the annotation task to gain deeper insights into these accuracy metrics. Our analysis reveals a notable imbalance in the distribution of "yes" labels, with high percentages observed for validity and contextualization (both at 93\%), in contrast to a lower percentage for constructiveness (39\%) and a moderate percentage for helpfulness (82\%).

To ensure a more balanced and thorough assessment of the model's evaluation capabilities across these dimensions, we have computed macro F1 scores. The results are as follows: validity at 83.38\%, contextualization at 70.24\%, constructiveness at 95.05\%, and helpfulness at 77.25\%. 

\subsection{Detailed Experiment Results}
\label{app:detailed_experiment_results}
\subsubsection{Results on Validity and Contextualization}
The results for validity and contextualization are in Tab.~\ref{tab:validity_context}. Our key findings are as follows: (i) Models typically produce valid and contextually appropriate feedback when no criteria or in-context demonstrations are given, or when only criteria are provided. (ii) For tasks with long in-context demonstrations (e.g., introduction and code tasks), adding in-context demonstrations will tend to distract some models with weak in-context ability (e.g., Together, WizCoder), leading to a noticeable decrease in contextually relevant feedback. Whereas if the in-context demonstrations are shorter (e.g., on reddit task), in-context demonstrations do not have a significant effect on the contextualization of the model-generated feedback. (iii) Among all models, the largest one (GPT-4) exhibits the highest validity and contextualization using different strategies to generate feedback, and the inclusion of criteria and in-context demonstrations do not distract the model much.

\begin{table}[t]
\centering
\small
  \setlength\tabcolsep{0.8pt}
  \renewcommand{\arraystretch}{1.1}
  \begin{tabular}{lcccccccc}
\toprule
\multicolumn{1}{l}{\multirow{2}{*}{\textbf{Intro.}}} 
& \multicolumn{4}{c}{\textbf{Validity}} & \multicolumn{4}{c}{\textbf{Contextualization}} \\
\cmidrule(lr){2-5} \cmidrule(lr){6-9}
\multicolumn{1}{c}{}                       & Base & Crit & ICL & CrICL & Base    & Crit   & ICL   & CrICL   \\
\midrule
Together  &  90.0 &   91.6   &  56.0   &   51.0      
&  80.0  & 84.8 & 37.6 &   26.0 \\
LAlpaca    &   96.0   &  97.8 & 95.0 &  93.8   &  93.0 & 96.0 & 78.8       &   74.6   \\
Command    &  98.0 & 90.0  & 95.0 &   87.4  &   96.0   &   82.6 &  59.6 &   57.0   \\
GPT4     &   100.0   &   98.2   & 94.2  &   96.8  &   100.0      &   96.8     &  94.2  &   94.8  \\
\midrule

\multicolumn{1}{l}{\multirow{2}{*}{\textbf{Reddit}}} 
& \multicolumn{4}{c}{\textbf{Validity}} & \multicolumn{4}{c}{\textbf{Contextualization}} \\
\cmidrule(lr){2-5} \cmidrule(lr){6-9}
\multicolumn{1}{c}{}                       & Base & Crit & ICL & CrICL & Base    & Crit   & ICL   & CrICL   \\
\midrule
Together    &  65.0 & 81.0 & 99.0  &  99.7    & 64.0  & 77.8  & 97.8 &  96.3    \\
LAlpaca   & 92.0 &  98.5 & 97.3 &   99.0  &  92.0 &  93.0  &  94.2 &   95.0  \\
Command   & 100.0 & 97.8 & 99.0 &  99.3   &  98.0  &    95.7 & 99.0   &   99.0    \\
GPT4     & 100.0 & 100.0&100.0 & 100.0   &  100.0  &  100.0 &  100.0 & 99.8    \\
\midrule

\multicolumn{1}{l}{\multirow{2}{*}{\textbf{Code}}} 
& \multicolumn{4}{c}{\textbf{Validity}} & \multicolumn{4}{c}{\textbf{Contextualization}} \\
\cmidrule(lr){2-5} \cmidrule(lr){6-9}
\multicolumn{1}{c}{}                       & Base & Crit & ICL & CrICL & Base    & Crit   & ICL   & CrICL   \\
\midrule
CLlama   & 100.0     & 99.6& 95.6   &   94.3  &  98.0 & 87.1 &  80.7& 72.6   \\
WizCoder   & 100.0 & 97.2 & 94.8 &   95.6  & 99.0 &81.6 & 54.5  &  64.2       \\
Claude2    & 100.0 &  99.7 & 99.2 &  97.4 &  100.0  &  87.3  & 89.7  &   82.2 \\
GPT4    &  100.0    & 99.7&  97.4 &  98.1   & 100.0   &  96.0 & 95.9  & 89.1  \\

\bottomrule
\end{tabular}
\caption{Performance of various LLMs on generating valid feedback and contextual feedback. ``Validity'' measures the percentage of feedback that is valid, and ``Contextualization'' measures the percentage of valid feedback that is contextual.}
\label{tab:validity_context}
\end{table}

\subsubsection{Additional Overall Performance}
We count the total number of criteria the generated feedback texts have touched upon by providing (not necessarily helpful) critiques or suggestions for the 100 test samples for each task. The results are shown in Tab.~\ref{tab:overall_performance_critique}. We observe that (i) Providing criteria only consistently helps models generate critiques or suggestions for more criteria. (ii) Providing demonstrations will generally encourage models to generate critiques or suggestions for more criteria except for the cases that the overly long demonstrations negatively affect the contextualization of feedback too much (in the case of introduction task for most models). (iii) Adding both criteria and demonstrations typically does not outperform adding criteria alone in terms of generating critiques and suggestions.

\begin{table}[t]
\centering
\small
  \setlength\tabcolsep{10pt}
  \renewcommand{\arraystretch}{0.8}
\begin{tabular}{lrrrr}
\toprule
\textbf{Intro.}   & Base & Crit & ICL  & CrICL \\
\midrule
Together & 250  & \textbf{369}  & 163  & 146   \\
LAlpaca  & 388  & \textbf{584}  & 225  & 190   \\
Command  & 374  & \textbf{487}  & 136  & 156   \\
GPT4     & 599  & \textbf{1025} & 706  & 704   \\
\midrule
\textbf{Reddit}   & Base & Crit & ICL  & CrICL \\
\midrule
Together & 209  & 295  & 316  & \textbf{343}   \\
LAlpaca  & 252  & \textbf{1003} & 326  & 361   \\
Command  & 355  & \textbf{1270} & 1020 & 1049  \\
GPT4     & 692  & \textbf{1437} & 1094 & 1110  \\
\midrule
\textbf{Code}     & Base & Crit & ICL  & CrICL \\
\midrule
CLlama   & 322  & \textbf{2095} & 966  & 947   \\
WizCoder & 420  & \textbf{1225} & 597  & 622   \\
Claude2  & 367  & \textbf{1493} & 1453 & 1086  \\
GPT4     & 511  & \textbf{1575} & 1081 & 1175 \\
\bottomrule
\end{tabular}
\caption{Overall performance of each model using different strategies in terms of the number of criteria the generated feedback texts touched upon through providing (not necessarily helpful) critiques or suggestions.}
\label{tab:overall_performance_critique}
\end{table}
\end{document}